\documentclass[11pt]{article}

\usepackage[preprint]{acl}
\usepackage{times}
\usepackage[T1]{fontenc}
\usepackage{latexsym}
\usepackage{microtype}
\usepackage{inconsolata}
\usepackage{graphicx}
\usepackage{booktabs}
\usepackage{makecell}
\usepackage{array}
\usepackage{tabularx}
\usepackage{pifont}
\usepackage{amsmath}
\usepackage{amssymb}
\usepackage{xcolor}
\usepackage{colortbl}
\usepackage[nointegrals]{wasysym}

\newcommand{\benchmarkname}{{\mdseries\textsc{FrameEvidence}}}

\title{\benchmarkname: A Benchmark for Reference Frames and Visual Evidence
in Multi-Image Spatial Reasoning}

\author{
  \normalfont Chunxi Luo$^{1,4,}$\thanks{Equal contribution.}\quad
  Qishuai Ding$^{1,*}$\quad
  Xingchen Peng$^{2,4,*}$\quad
  Hengwei Ye$^{4}$\quad
  ke zhang$^{4}$\\
  \normalfont Wenxin Lian$^{1}$\quad
  Tianyi Lu$^{3,4}$\quad
  Chunlong Yu$^{2,4}$\quad
  Zhenya Lin$^{4,}$\thanks{Corresponding author.}\\[4pt]
  {\small\normalfont $^{1}$Shanghai University\quad
    $^{2}$East China University of Science and Technology}\\
  {\small\normalfont $^{3}$Shanghai Ocean University\quad
    $^{4}$Byaidata}\\[3pt]
  {\small\normalfont Chunxi Luo\quad
    \href{mailto:lcxshu@163.com}{\texttt{lcxshu@163.com}}}
}

\hypersetup{
  pdftitle={FrameEvidence: A Benchmark for Reference Frames and Visual Evidence in Multi-Image Spatial Reasoning},
  pdfauthor={Chunxi Luo, Qishuai Ding, Xingchen Peng, Hengwei Ye, ke zhang, Wenxin Lian, Tianyi Lu, Chunlong Yu, Zhenya Lin},
  pdfsubject={Multi-image spatial reasoning benchmark},
  pdfkeywords={spatial reasoning, reference frames, visual evidence, multi-image reasoning, benchmark}
}

\begin{document}
\flushbottom
\maketitle

\begin{abstract}
A spatial answer depends on the reference frame specified in language
and its supporting visual evidence. Item accuracy alone obscures
whether correctness persists when either changes. We introduce
\benchmarkname{}, a benchmark of 320 matched triplets and 960 Chinese
multiple-choice questions from captured and rendered scenes. One family
compares three grounded reference frames over fixed images. The other
pairs image deletion annotated as insufficient with answer-preserving addition.
Joint metrics require correct answers across all three frames and paired
evidence variants. Across 20 configurations
on 288 held-out questions, GPT three-shot reaches 50.00\% item accuracy,
but the best three-frame group accuracy is 5/48 (10.42\%), reached by GPT
two-shot. Few-shot prompting and task-specific adaptation improve item
accuracy without consistently improving joint correctness. Input controls
show where images support matched success, while repeated executions reveal
answer instability beneath similar scores.
\end{abstract}

\section{Introduction}
\label{sec:introduction}

Spatial language describes an arrangement relative to a reference frame.
An object on the left from a camera may be in front in another object's frame.
Models must identify the intended frame and assess whether partial views
jointly support that description. Spatial supervision has improved
spatial question answering and regional reasoning in vision-language models
\citep{Chen_2024_CVPR,Cheng_2024_NeurIPS}. Evaluation must therefore test
both frame grounding and visual evidence use.

Existing benchmarks test perceptual and three-dimensional spatial relations
\citep{liu2023visual,fu2024blink,Ma_2025_ICCV}, as well as reference-frame ambiguity
and subject-centered judgments
\citep{ambiguitiesvision,premsri2025forest,liu2026mirrorqa}.
Multi-view and video evaluations extend this coverage to evidence distributed
across observations \citep{yang2026mmsi,yeh2026seeing,Yang_2025_CVPR}.
Yet item accuracy alone cannot establish correctness across matched inputs.

A preferred answer can match a direction without reflecting the specified
frame. Matched evaluations distinguish these behaviors by changing selected
input attributes
\citep{thrush2022winoground,wang2025muirbench,kim2026diagnosing}.
The key is whether the correct answer should \emph{change or remain fixed}.

\begin{figure*}[t]
  \centering
  \includegraphics[width=\textwidth]{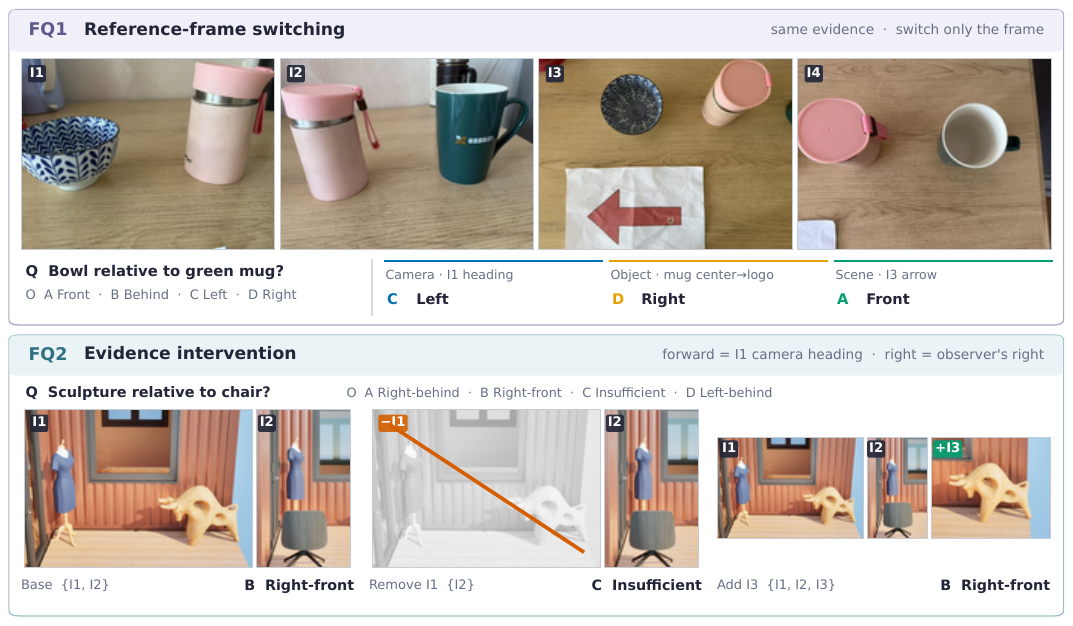}
  \caption{Matched protocols. FQ1 changes the queried frame while fixing
  the images. FQ2 deletes an annotated required view or adds a hard non-MSS
  view. The crossed-out image is withheld, and I1--I4 mark original positions
  for display. After I1 is deleted, ``the first image'' refers to I2.}
  \label{fig:task-design}
\end{figure*}

Figure~\ref{fig:task-design} presents FrameEvidence's two question families.
The first (FQ1) fixes ordered images and the
target--relatum pair while querying camera, object-anchor and scene-axis frames.
The second (FQ2) pairs an annotated minimal sufficient set (MSS) with deletion
and addition variants, labeled as insufficient and answer-preserving, respectively.
We also analyze changes to image-order references after deletion.
These comparisons extend rationale, unanswerability and view-selection
diagnostics of evidence dependence
\citep{deyoung2020eraser,moratelli2026benchmarking,guo2026ssmnbench,
wang2026does}.
FQ1 joint scoring requires three correct frame answers. FQ2 separately
requires a correct MSS answer together with a correct deletion or addition answer.
Response transitions then distinguish initial errors from failures after the change.

The release contains 320 triplets and 960 questions, with 96 triplets and
288 questions reserved for testing. We evaluate ten open-weight checkpoints,
two adapted counterparts and two gateway aliases under zero- to three-shot prompting.
The strongest item-level setting answers half the test correctly but solves
only three of 48 frame groups. Images improve MSS--addition success for
eleven of twelve open-weight configurations, but item accuracy for only six.
We compare low-rank adaptation (LoRA) across three seeds and pair
option rotations with same-order repeats to separate order sensitivity from
execution variability.


\section{Related Work}
\label{sec:related-work}

\begin{table*}[t]
  \centering
  \setlength{\abovecaptionskip}{0pt}
  \setlength{\belowcaptionskip}{5pt}
  \caption{Comparison of evaluation protocols most relevant to FrameEvidence.}
\label{tab:protocol-comparison}

  \begingroup
  \def\FEyesmark{\textbf{Y}}
  \def\FEpartialmark{\textit{P}}
  \def\FEnomark{--}
  \let\FEtoprule\toprule
  \let\FEmidrule\midrule
  \let\FEbottomrule\bottomrule
  \footnotesize
  \setlength{\tabcolsep}{2.2pt}
  \renewcommand{\arraystretch}{0.94}
  \begin{tabular*}{\textwidth}{@{\extracolsep{\fill}}lccccccccl@{}}
    \FEtoprule
    Benchmark & Src. & MI & Match & \shortstack{FoR\\S} &
    \shortstack{Min\\Ev} & \shortstack{Del.\\$\to\bot$} &
    Hard$+$ & Joint & Scale \\
    \FEmidrule
    COMFORT~\citep{ambiguitiesvision}
      & Syn. & \FEnomark & \FEyesmark & \FEyesmark & \FEnomark & \FEnomark & \FEnomark & \FEpartialmark
      & 58.3K English cases \\
    MuirBench~\citep{wang2025muirbench}
      & Mix. & \FEyesmark & \FEyesmark & \FEnomark & \FEnomark & \FEnomark & \FEnomark & \FEpartialmark
      & 2,600 instances \\
    MMSI-Bench~\citep{yang2026mmsi}
      & Real & \FEyesmark & \FEnomark & \FEnomark & \FEpartialmark & \FEnomark & \FEnomark & \FEnomark
      & 1,000 QA \\
    All-Angles~\citep{yeh2026seeing}
      & Real & \FEyesmark & \FEyesmark & \FEpartialmark & \FEnomark & \FEnomark & \FEnomark & \FEyesmark
      & 2,132 QA \\
    SCOPE~\citep{kim2026diagnosing}
      & Mix. & \FEyesmark & \FEyesmark & \FEpartialmark & \FEnomark & \FEnomark & \FEnomark & \FEpartialmark
      & 20,144 QA \\
    MultiView-Bench~\citep{zhang2026multiview}
      & Syn. & \FEyesmark & \FEpartialmark & \FEnomark & \FEnomark & \FEnomark & \FEnomark & \FEnomark
      & 500 main ({+}2K var.) \\
    SSMNBench~\citep{guo2026ssmnbench}
      & Real & \FEyesmark & \FEyesmark & \FEnomark & \FEyesmark & \FEnomark & \FEpartialmark & \FEnomark
      & 3,300 QA \\
    View-Level Evidence~\citep{wang2026does}
      & Real & \FEyesmark & \FEpartialmark & \FEnomark & \FEpartialmark & \FEnomark & \FEnomark & \FEpartialmark
      & 122 QA \\
    \FEmidrule
    \textbf{FrameEvidence}
      & \textbf{Mix.} & \FEyesmark & \FEyesmark & \FEyesmark
      & \FEyesmark & \FEyesmark & \FEyesmark & \FEyesmark
      & \textbf{960 QA (320 triplets)} \\
    \FEbottomrule
  \end{tabular*}
  \vspace{1pt}
  \parbox{\textwidth}{\footnotesize\raggedright
    \textit{Legend.} \FEyesmark/\FEpartialmark/\FEnomark\ denote explicit,
    partial, and unspecified coverage in the reported protocols.
    MI, Match, FoR-S, and Min-Ev abbreviate multi-image, item matching,
    frame switching, and minimal evidence. Del.$\to\bot$ denotes deletion to
    insufficiency, Hard$+$ denotes truth-preserving hard non-MSS addition, and
    Joint denotes correctness across matched variants. Partial Joint coverage
    includes response consistency or joint view--answer prediction.
    Scales use source units.}
  \endgroup

\end{table*}

\paragraph{Reference frames and frame-conditioned reasoning.}
Spatial and matched-contrast benchmarks probe relational judgments
\citep{liu2023visual,fu2024blink,thrush2022winoground}.
COMFORT, FoREST, and ViewSpatial-Bench examine reference-frame ambiguity
\citep{ambiguitiesvision,premsri2025forest,li2026viewspatial}.
SCOPE, All-Angles, and CityCube address cross-view and scale variation
\citep{kim2026diagnosing,yeh2026seeing,xu2026citycube}, while MirrorQA targets
subject-centered left--right judgments \citep{liu2026mirrorqa}. COMFORT
compares explicit frames on matched synthetic scenes and measures
probabilistic consistency. Our frame protocol evaluates joint correctness
across three multi-view frame prompts, distinguishing a correct switch from
an answer change alone.

\paragraph{Multi-image reasoning and evidence interventions.}
MIRB, ReMI, MMIU, and MuirBench assess multi-image reasoning, while MIMIC
probes controlled failure cases
\citep{zhao2024benchmarking,kazemi2024remi,meng2025mmiu,wang2025muirbench,
das2026more}. MMSI-Bench, CrossView Suite, MultiView-Bench, and cross-view
relation learning focus on spatial integration
\citep{yang2026mmsi,wang2026crossview,zhang2026multiview,
jeong2026learning}. Evidence-centered diagnostics examine input dependence through
rationale sufficiency, paired unanswerability, view selection, evidence
sensitivity, and multi-view necessity
\citep{deyoung2020eraser,moratelli2026benchmarking,wang2025muirbench,
guo2026ssmnbench,wang2026does,bhat2026consistent,yu2026mindedit}.
SSMNBench adds and removes views of minimal necessary sets to measure
view-wise accuracy and distraction decay. These measures characterize supporting
views and average sensitivity to evidence changes. Table~\ref{tab:protocol-comparison}
summarizes our requirement for paired correctness before and after an
intervention on the same instance. Comparing models on
shared, correctly answered MSSs then separates initial coverage from
subsequent intervention errors.

\section{Problem Formulation}
\label{sec:problem-formulation}

Each item is $x=(\mathbf{I},q,\mathcal{O})$, where
$\mathbf{I}=(I_1,\ldots,I_m)$ contains one to five images, $q$ is a Chinese
question, and $\mathcal{O}=(o_A,o_B,o_C,o_D)$ is a four-option answer set.
A model returns $\hat y=h_\theta(\mathbf{I},q,\mathcal{O})\in\{A,B,C,D\}$.
Spatial answers use axial (front, back, left, right) or quadrant directions
(front-left, front-right, back-left, back-right). FQ2 includes insufficient
information. Metric reconstruction, vertical relations and motion are excluded.

\paragraph{Reference-frame intervention (FQ1).}
Each group queries the same target--relatum relation in three frames
$\mathcal{F}=\{\mathrm{cam},\mathrm{obj},\mathrm{scene}\}$. The camera frame
uses a specified camera heading, the object frame a named object's orientation,
and the scene frame an environmental axis. The question specifies the anchor,
which need not be the relatum, and its forward or backward convention.
Ordered images, entity identities, relation ontology and option meanings
remain fixed. Only the frame instruction and its implied answer change.
At least two gold answers differ, so repeating one answer cannot solve the group.

\paragraph{Evidence intervention (FQ2).}
The base set $M_g$ is an annotated MSS supporting spatial answer $y_g$
when no proper subset does.
Deletion removes an annotated required image $r_g$. Addition restores the
MSS and adds a hard, truth-preserving image $d_g\notin M_g$.
The annotated responses to $M_g$,
$M_g\setminus\{r_g\}$ and $M_g\cup\{d_g\}$ are respectively $y_g$,
$\bot$ and $y_g$, where $\bot$ means insufficient information.
This definition requires subset-minimality, not minimum cardinality.
The target--relatum text stays fixed, but deletion can change image-ordinal
bindings, as analyzed in Section~\ref{sec:dataset-analysis}.

\paragraph{Group-level evaluation.}
Let $c_g^v=\mathbb{1}[\hat y_g^v=y_g^v]$ denote correctness for variant $v$,
and $n_k=|\mathcal{G}_k|$ the number of groups in family $k$.
FQ1 Group Exact Match scores the three frame answers jointly,
\begin{equation}
\label{eq:group-success}
\mathrm{FQ1\text{-}GEM}
 =n_1^{-1}\sum_{g\in\mathcal{G}_1}\prod_{f\in\mathcal{F}}c_g^f.
\end{equation}
For FQ2, Necessity Transition Success is
$\mathrm{NTS}=n_2^{-1}\sum_g c_g^M c_g^{-}$, and Distractor Invariance
Success is $\mathrm{DIS}=n_2^{-1}\sum_g c_g^M c_g^{+}$, with both sums over
$\mathcal{G}_2$. NTS scores agreement with both the MSS and deletion labels,
while DIS requires the spatial answer to remain correct after addition.

\begin{table*}[t]
  \begingroup
  \def\FEpairedtables{1}
  \begin{minipage}[t]{\columnwidth}
    \vspace{0pt}
    \ifdefined\FEpairedtables\else\begin{table}[t]\fi
  \centering
  \setlength{\abovecaptionskip}{0pt}
  \setlength{\belowcaptionskip}{5pt}
  \caption{Dataset composition by split. All entries count questions except the Groups row.}
  \label{tab:dataset-statistics}
  \begingroup\footnotesize
  \setlength{\tabcolsep}{3pt}
  \renewcommand{\arraystretch}{1.07}
  \begin{tabular*}{\columnwidth}{@{\extracolsep{\fill}}lrrrr@{}}
    \toprule
    Category & Train & Dev & Test & Total \\
    \midrule
    Groups & 192 & 32 & 96 & 320 \\
    Questions & 576 & 96 & 288 & 960 \\
    \midrule
    FQ1 & 288 & 48 & 144 & 480 \\
    FQ2 & 288 & 48 & 144 & 480 \\
    \midrule
    Tabletop & 195 & 27 & 102 & 324 \\
    Indoor large & 192 & 33 & 93 & 318 \\
    Outdoor public & 189 & 36 & 93 & 318 \\
    \midrule
    Axial & 291 & 45 & 144 & 480 \\
    Quadrant & 285 & 51 & 144 & 480 \\
    \bottomrule
  \end{tabular*}
  \endgroup
\ifdefined\FEpairedtables\else\end{table}\fi

  \end{minipage}\hfill
  \begin{minipage}[t]{\columnwidth}
    \vspace{0pt}
    \ifdefined\FEpairedtables\else\begin{figure}[t]\fi
  \begingroup
  \makeatletter\def\@captype{figure}\makeatother
  \centering
  \includegraphics[width=\columnwidth]{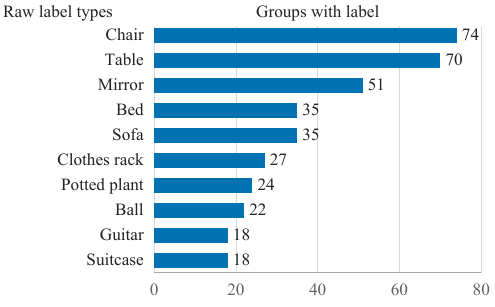}
  \caption{Object-label counts per group. The other 116 labels contribute
  586 of the 960 label occurrences.}
  \label{fig:object-vocabulary-main}
  \endgroup
\ifdefined\FEpairedtables\else\end{figure}\fi

  \end{minipage}
  \endgroup
\end{table*}

\section{FrameEvidence Construction}
\label{sec:construction}

\subsection{Scene Acquisition and Group Assembly}
\label{sec:image-collection}

FrameEvidence combines 126 captured-image and 194 rendered-image groups of
static layouts spanning indoor tabletop, indoor large-space and outdoor
public-space scenes. An object inventory identifies the target, relatum
and any bridge objects connecting views. Construction criteria require
unambiguous identity matches, unchanged layouts, grounded frame anchors
and evidence that must be combined across views.

FQ1 reuses two to four ordered images across its three questions.
FQ2 starts from two to four MSS images, then changes the set size by one.
Each family contains 53 two-image, 54 three-image and 53 four-image base sets.
Hard additions supply competing context but preserve the annotated answer.

\subsection{Annotation and Quality Control}
\label{sec:annotation-guidelines}

Controlled Chinese questions specify the queried entities and reference frame.
Answer placement maps semantic answers to A--D while fixing option order
within groups. Each letter occurs 240 times overall and 40 per condition.

Each group has approval from a separate reviewer account with answers visible.
The checklist covers scenes, bindings, directions, each MSS image's necessity
and answer preservation. Appendix~\ref{sec:production-qc} documents this review
and the automated schema and consistency checks.

\subsection{Split and Validation Scope}
\label{sec:split-human-validation}

Table~\ref{tab:dataset-statistics} allocates 192/32/96 complete groups to
training/development/test. Matched variants and identical images stay
together, keeping triplets intact across splits.
Different views can share a scene across splits. Evaluation therefore
concerns held-out groups, not necessarily unseen scenes.

Table~\ref{tab:main-results} reports one participant's sampled blind-answer
summary. Appendix~\ref{sec:implementation} describes this reference and the
exact random baseline.

\section{FrameEvidence Analysis}
\label{sec:dataset-analysis}

\begin{figure*}[t]
  \centering
  \includegraphics[width=\textwidth]{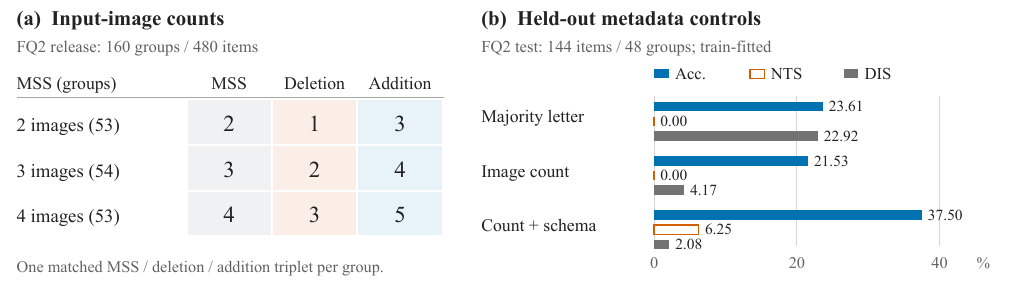}
  \caption{FQ2 image counts across the release (a) and train-fitted metadata
  controls (b). Test accuracy uses 144 items and NTS/DIS 48 groups.
  Schema is the visible insufficient-information option's letter.}
  \label{fig:evidence-audit}
\end{figure*}

\paragraph{Composition and balance.}
Table~\ref{tab:dataset-statistics} shows balanced relation ontologies and
nearly equal scene strata, with 160 questions per condition.
The ten labels in Figure~\ref{fig:object-vocabulary-main} account for 374 of
960 group-level occurrences, with 116 further types retaining source names.
Test A--D counts are 75, 70, 69 and 74. Macro-averaging gives each class
equal weight.

\paragraph{Evidence structure and non-visual controls.}
Figure~\ref{fig:evidence-audit}a shows that extreme image counts identify
some FQ2 conditions. On 144 held-out FQ2 items, a train-fitted
count-plus-schema rule achieves 37.50\% accuracy, compared with 33.33\%
from always selecting insufficient information. NTS and DIS reach only
6.25\% and 2.08\%, so this advantage rarely extends to paired variants.
A classifier using character-level term frequency--inverse document
frequency (TF-IDF) features from questions and options reaches 23.96\%
accuracy on all 288 test items. Identical FQ2 text forces this deterministic
model to repeat its answer, making NTS zero. Condition cues and item accuracy
alone therefore cannot establish sensitivity to changed evidence.
Appendix~\ref{sec:offline-text} specifies the text model.

\paragraph{Scene overlap and sensitivity.}
The release has 443 distinct images across 1,120 paths. Exact images stay
within one split, although views and renderings can share scenes across splits.
An inferred scene/template ledger assigns 81 test groups to clusters
shared with training and 15 to four clusters absent from both train and dev.
On these 15 groups, Qwen LoRA lowers DIS from 4/8 to 2/8, matching the
full-test direction. InternVL's unchanged overall NTS instead combines
opposing changes across strata. The subset's 13 rendered and two captured
groups also differ in composition from the full test.
Across the full test, rendered inputs yield higher DIS in 16/20
configurations and higher accuracy in 14. Appendices~\ref{sec:scene-overlap-supplement}
and~\ref{sec:source-crossrun} report both comparisons.

\paragraph{Ordinal references after deletion.}
Deletion removes the original referent in 36/48 FQ2 groups and shifts it in six.
Under current-order presentation, 38 ordinals resolve to another image and
four fall out of range. Table~\ref{tab:deletion-reference-main} separates joint
successes by reference change. Both InternVL checkpoints succeed only when
the referent is removed. Their NTS scores therefore combine responses to
evidence removal and reference changes.

\begin{table}[t]
  \centering
  \setlength{\abovecaptionskip}{0pt}
  \setlength{\belowcaptionskip}{4pt}
  \caption{Joint deletion successes by reference change.}
  \label{tab:deletion-reference-main}
  \begingroup\fontsize{9}{10}\selectfont
  \setlength{\tabcolsep}{3pt}
  \begin{tabular*}{\columnwidth}{@{\extracolsep{\fill}}lrrrr@{}}
    \toprule
    Model & Removed & Shifted & Stable & None \\
    Groups & 36 & 6 & 2 & 4 \\
    \midrule
    Qwen Base & 1 & 0 & 0 & 0 \\
    Qwen LoRA & 5 & 0 & 1 & 0 \\
    InternVL Base & 4 & 0 & 0 & 0 \\
    InternVL LoRA & 4 & 0 & 0 & 0 \\
    \midrule[0.3pt]
    GPT 2-shot & 8 & 3 & 0 & 0 \\
    GPT 3-shot & 7 & 2 & 1 & 0 \\
    Claude 0-shot & 4 & 0 & 0 & 0 \\
    Claude 3-shot & 3 & 1 & 0 & 0 \\
    \bottomrule
  \end{tabular*}
  \par\vspace{2pt}
  \parbox{\columnwidth}{\footnotesize\raggedright
  Joint MSS--deletion counts. Groups gives NTS denominators. Qwen/InternVL
  are primary 7B/8B checkpoints. None denotes no ordinal.
  Full results appear in Appendix~\ref{sec:ordinal-audit}.}
  \endgroup
\end{table}

Camera headings also occur in 30/48 MSS prompts with non-camera tags.
These tags define annotation strata, not distinct frame conditions.

\section{Experiments}
\label{sec:experiments}

\begin{table*}[t]
  \centering
  \setlength{\abovecaptionskip}{0pt}
  \setlength{\belowcaptionskip}{5pt}
  \caption{All-images model results (\%) on 288 questions, with 48 groups per
  family for joint metrics.}
  \label{tab:main-results}

  \begingroup\footnotesize
  \setlength{\tabcolsep}{3pt}
  \begin{tabular*}{\textwidth}{@{\extracolsep{\fill}}llrrrrrrr@{}}
    \toprule
    & & \multicolumn{4}{c}{Classification} & \multicolumn{3}{c}{Matched success} \\
    \cmidrule(lr){3-6}\cmidrule(l){7-9}
    Model & Setting & P & R & F1 & Acc. & GEM & NTS & DIS \\
    \midrule
    \rowcolor{blue!5}\multicolumn{9}{c}{\textbf{\textit{Open-weight baselines}}} \\
    Qwen2.5-VL-7B & Base & 31.06 & 29.43 & 25.99 & 28.82 & 0.00 & 2.08 & 43.75 \\
    InternVL3-8B & Base & 29.72 & 29.84 & 28.65 & 29.51 & 2.08 & 8.33 & 20.83 \\
    Qwen3-VL-8B & Base & 34.60 & 34.30 & 33.83 & 34.03 & 0.00 & 14.58 & 39.58 \\
    LLaVA-OneVision-7B & Base & 30.56 & 29.33 & 29.04 & 29.17 & 0.00 & 0.00 & 37.50 \\
    MiniCPM-V 4.5 & Base & 33.48 & 33.33 & 32.27 & 32.99 & 0.00 & 4.17 & \textbf{47.92} \\
    Mistral Small 3.2 24B & Base & 26.99 & 25.96 & 24.82 & 25.69 & 0.00 & 2.08 & 8.33 \\
    Qwen2.5-VL-32B & Base & 33.29 & 33.56 & 32.15 & 33.33 & 0.00 & 6.25 & 35.42 \\
    InternVL3.5-14B & Base & 28.14 & 28.35 & 27.86 & 28.12 & 0.00 & 6.25 & 25.00 \\
    Phi-4-Multimodal & Base & 26.07 & 26.61 & 25.54 & 26.39 & 0.00 & 0.00 & 35.42 \\
    Kimi-VL-A3B & Base & 36.74 & 29.48 & 26.73 & 29.17 & 0.00 & 0.00 & 39.58 \\
    \midrule
    \rowcolor{blue!5}\multicolumn{9}{c}{\textbf{\textit{Task-adapted counterparts}}} \\
    Qwen2.5-VL-7B & LoRA & 35.45 & 35.53 & 34.81 & 35.42 & 0.00 & 12.50 & 12.50 \\
    InternVL3-8B & LoRA & 34.35 & 32.91 & 30.99 & 32.64 & 0.00 & 8.33 & 8.33 \\
    \midrule
    \rowcolor{blue!5}\multicolumn{9}{c}{\textbf{\textit{Gateway-served models}}} \\
    GPT-5.6 & 0-shot & 40.57 & 40.72 & 40.43 & 40.62 & 4.17 & 14.58 & 35.42 \\
     & 1-shot & 44.01 & 44.26 & 44.05 & 44.10 & 6.25 & 16.67 & 41.67 \\
     & 2-shot & 48.53 & 48.29 & 48.34 & 48.26 & \textbf{10.42} & \textbf{22.92} & 43.75 \\
     & 3-shot & \textbf{49.87} & \textbf{50.05} & \textbf{49.86} & \textbf{50.00} & 6.25 & 20.83 & \textbf{47.92} \\
    Claude Fable 5.1 & 0-shot & 39.35 & 39.53 & 39.19 & 39.24 & 0.00 & 8.33 & 41.67 \\
     & 1-shot & 40.55 & 40.30 & 40.26 & 40.28 & 6.25 & 2.08 & 37.50 \\
     & 2-shot & 41.34 & 41.58 & 41.27 & 41.32 & 6.25 & 12.50 & 37.50 \\
     & 3-shot & 41.75 & 41.76 & 41.64 & 41.67 & 2.08 & 8.33 & 39.58 \\
    \midrule
    \rowcolor{gray!8}\multicolumn{9}{c}{\textit{Random and human references}} \\
    Uniform random & Expected & 25.00 & 25.00 & 24.93 & 25.00 & 1.56 & 6.25 & 6.25 \\
    Human (reported) & Sampled & 98.56 & 98.40 & 98.48 & 98.40 & -- & -- & -- \\
    \bottomrule
  \end{tabular*}
  \par\smallskip\parbox{\textwidth}{\footnotesize Model P/R/F1 are macro-averaged over A--D.
  GEM denotes FQ1-GEM. Bold marks model-column maxima.
  LoRA uses one checkpoint per family. Invalid answers count as incorrect.
  Random gives finite-test expectations for independent uniform A--D guesses.
  Human reports one participant's sampled blind evaluation, as summarized
  by the authors. Dashes denote unreported joint scores.
  Appendix~\ref{sec:implementation} details the reference rows.}
  \endgroup
  
\end{table*}

\subsection{Evaluation Setup}

Table~\ref{tab:main-results} compares 20 configurations on 288 questions.
GPT-5.6 and Claude Fable 5.1 denote the requested gateway aliases.
The main evaluation and input controls share checkpoints. Table~\ref{tab:ordinary-multiseed-main}
compares six adapters with separate Base runs, which agree with primary runs on
281/288 Qwen and 277/288 InternVL answers.
Appendix~\ref{sec:source-crossrun} reports both paired comparisons.

Queries preserve image order and request a single letter. Open-weight and Claude
runs use temperature zero and 32 output tokens. GPT instead leaves
temperature unset and allows 2,048 output tokens with reasoning effort
set to low.

Following MirrorQA \citep{liu2026mirrorqa}, we evaluate zero- to three-shot
prompting. Demonstrations use nested prefixes of one training group per
family, selected with seed 20260722, excluding test groups and exact test images.

We report accuracy and macro-averaged precision, recall and F1 over A--D on all
288 questions, with joint metrics over each family's 48 groups.
Ten LLaVA-OneVision responses are invalid and count as errors.
All other main responses are valid.

\ifdefined\FEpairedtables\else\begin{table}[!t]\fi
  \centering
  \setlength{\abovecaptionskip}{0pt}
  \setlength{\belowcaptionskip}{5pt}
  \caption{All-images minus text-only in percentage points,
  with questions and gold labels fixed.}
  \label{tab:input-dependence-all-main}
  \begingroup\footnotesize
  \setlength{\tabcolsep}{1.5pt}
  \renewcommand{\arraystretch}{1.12}
  \begin{tabular*}{\columnwidth}{@{\extracolsep{\fill}}lrrr@{}}
    \toprule
    Model & $\Delta$Acc. & $\Delta$NTS & $\Delta$DIS \\
    \midrule
    Qwen2.5-VL-7B & $-3.1$ & $+2.1$ & $+43.8$ \\
    Qwen2.5-VL-7B LoRA & $+3.1$ & $+12.5$ & $+12.5$ \\
    InternVL3-8B & $-1.4$ & $+8.3$ & $+12.5$ \\
    InternVL3-8B LoRA & $+2.4$ & $+8.3$ & $+8.3$ \\
    Qwen3-VL-8B & $+4.5$ & $+14.6$ & $+33.3$ \\
    LLaVA-OneVision-7B & $+0.0$ & $+0.0$ & $-6.2$ \\
    MiniCPM-V 4.5 & $+3.8$ & $+4.2$ & $+14.6$ \\
    Mistral Small 3.2 24B & $-3.8$ & $+2.1$ & $+8.3$ \\
    Qwen2.5-VL-32B & $+2.4$ & $+6.2$ & $+35.4$ \\
    InternVL3.5-14B & $-2.1$ & $+6.2$ & $+10.4$ \\
    Phi-4-Multimodal & $-2.4$ & $+0.0$ & $+16.7$ \\
    Kimi-VL-A3B & $+4.2$ & $+0.0$ & $+4.2$ \\
    \midrule
    GPT-5.6 / 0-shot & $\mathbf{+8.7}$ & $+14.6$ & $+35.4$ \\
    \bottomrule
  \end{tabular*}
  \par\smallskip
  \parbox{\columnwidth}{\footnotesize\raggedright
  Positive values favor all-images. Unmarked rows are Base, and LoRA uses
  primary adapters. Each comparison covers 288 questions and 48 groups per
  joint metric. Appendix~\ref{sec:stratified-results} gives paired intervals.}
  \endgroup
\ifdefined\FEpairedtables\else
\end{table}\fi

\subsection{Item Accuracy and Matched Success}

Item accuracy and joint scores identify different leaders in
Table~\ref{tab:main-results}. GPT three-shot reaches \textbf{50.00\%} accuracy,
while two-shot leads
FQ1-GEM and NTS at 10.42\% and \textbf{22.92\%}.
MiniCPM-V-4.5 matches three-shot's \textbf{47.92\% DIS} at 32.99\% accuracy.
GPT's 1.74-point two-to-three-shot gain combines 48 newly correct and 43
newly incorrect answers, masking substantial turnover. Its paired 95\%
interval [$-4.90$, $5.35$] includes zero.
Appendix~\ref{sec:shot-uncertainty} compares all adjacent shot settings.

This divergence depends on how errors align across variants. The product of
variant accuracies gives the expected joint rate under independent
cross-group rearrangement. NTS falls below this reference in 19/20
configurations, while DIS exceeds it in all 20. An image-free majority
control yields the same asymmetry because repeating an answer can succeed
on answer-preserving addition but miss deletion's insufficiency label.
Appendix~\ref{sec:joint-marginals} details \mbox{this shared-answer contribution}.

\begin{table*}[t]
  \centering
  \setlength{\abovecaptionskip}{0pt}
  \setlength{\belowcaptionskip}{5pt}
  \caption{Ordinary and Circular evaluation (\%) on 288 questions.
  Open-weight controls are matched within panels.
  LoRA gives three-seed mean $\pm$ sample standard deviation. GPT is zero-shot.}
  \label{tab:adaptation-summary}
  \label{tab:ordinary-multiseed-main}
  \label{tab:circular-summary}
  \begin{minipage}[t]{\columnwidth}
    \centering\footnotesize
    \textbf{(a) Ordinary all-images evaluation}\par\smallskip
    \setlength{\tabcolsep}{1.4pt}
    \begin{tabular*}{\columnwidth}{@{\extracolsep{\fill}}lrrrr@{}}
      \toprule
      Setting & Acc. & GEM & NTS & DIS \\
      \midrule
      \multicolumn{5}{@{}l}{\textit{InternVL3-8B}} \\
      Base & 30.21 & 2.08 & 8.33 & 20.83 \\
      LoRA & $32.41{\pm}1.31$ & $0.69{\pm}1.20$ & $9.72{\pm}1.20$ & $12.50{\pm}2.08$ \\
      \midrule
      \multicolumn{5}{@{}l}{\textit{Qwen2.5-VL-7B}} \\
      Base & 29.17 & 0.00 & 2.08 & 45.83 \\
      LoRA & $36.00{\pm}1.40$ & $0.00{\pm}0.00$ & $10.42{\pm}3.61$ & $12.50{\pm}0.00$ \\
      \midrule
      \multicolumn{5}{@{}l}{\textit{GPT-5.6}} \\
      0-shot & 40.62 & 4.17 & 14.58 & 35.42 \\
      \bottomrule
    \end{tabular*}
  \end{minipage}\hfill
  \begin{minipage}[t]{\columnwidth}
    \centering\footnotesize
    \textbf{(b) Four option rotations}\par\smallskip
    \setlength{\tabcolsep}{2pt}
    \begin{tabular*}{\columnwidth}{@{\extracolsep{\fill}}lrrr@{}}
      \toprule
      Setting & Vanilla & Circular & Consistency \\
      \midrule
      \multicolumn{4}{@{}l}{\textit{InternVL3-8B}} \\
      Base & 30.21 & 7.64 & 23.61 \\
      LoRA & $32.75{\pm}1.31$ & $6.02{\pm}1.12$ & $17.59{\pm}1.57$ \\
      \midrule
      \multicolumn{4}{@{}l}{\textit{Qwen2.5-VL-7B}} \\
      Base (S) & 14.93 & 2.08 & 5.90 \\
      LoRA & $35.19{\pm}1.12$ & \textbf{16.20}$\pm$\textbf{0.53} & $32.64{\pm}1.25$ \\
      \midrule
      \multicolumn{4}{@{}l}{\textit{GPT-5.6}} \\
      0-shot & 45.49 & \textbf{18.75} & 27.08 \\
      \bottomrule
    \end{tabular*}
  \end{minipage}
  \par\smallskip\parbox{\textwidth}{\footnotesize
  GEM is FQ1-GEM. Joint scores use 48 groups per family. Vanilla uses within-run
  predictions. Appendix~\ref{sec:implementation} gives the source of the
  summary marked S and all per-seed metrics.}
\end{table*}

\begin{figure*}[t]
  \centering
  \includegraphics[width=\textwidth]{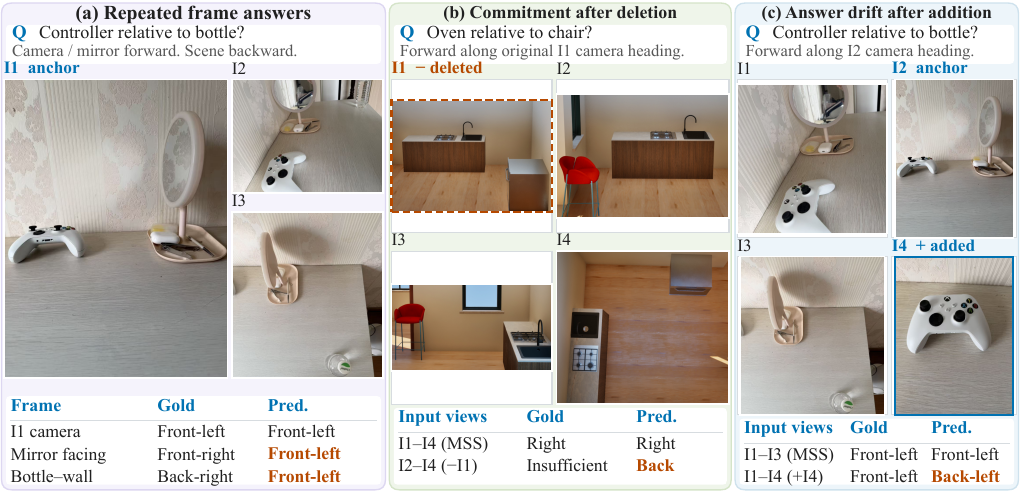}
  \caption{Selected GPT three-shot errors with released gold labels.
  I1--I4 mark original positions for display. Inputs use unlabeled images
  in their current order. Deletion in (b) removes the original camera anchor.}
  \label{fig:typical-errors}
\end{figure*}

\subsection{Input Controls and Adaptation}

For twelve open-weight configurations and GPT zero-shot, text-only removes
all images and single-image keeps the first. Questions, options and labels
stay fixed in Table~\ref{tab:input-ablation}.

\begin{figure*}[t]
  \centering
  \includegraphics[width=\textwidth]{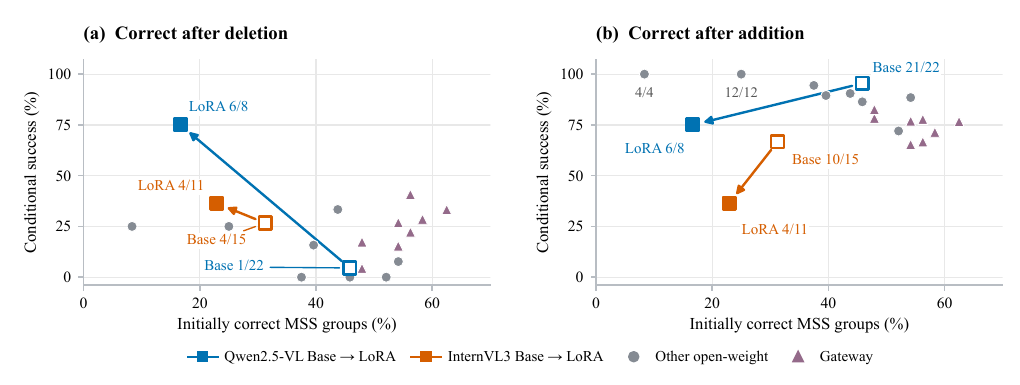}
  \caption{Initial and conditional success across 20 primary configurations.
  Horizontal rates use all 48 FQ2 groups. Vertical rates use each
  configuration's MSS-correct subset, with counts on highlighted points.
  Arrows link Base and LoRA, each with its own conditioning subset.}
  \label{fig:conditional-interventions}
\end{figure*}

Visual input improves addition-pair success more often than item accuracy.
In Table~\ref{tab:input-dependence-all-main}, images raise DIS for 11/12
open-weight configurations but accuracy for only six. Qwen2.5-VL-7B gains
\textbf{43.8 DIS points}, with a 95\% interval of [29.2, 59.5], despite lower
accuracy. GPT gains \textbf{8.7~accuracy~points} over text-only.
Without images, it always selects insufficiency for FQ2. Using all images
instead of the first raises its DIS from 8.33\% to 35.42\%.
Figure~\ref{fig:conditional-interventions} separates initial MSS correctness
from post-addition answer retention, both of which \mbox{contribute to DIS}.

Adaptation improves item accuracy without preserving addition-pair success.
Across three seeds in
Table~\ref{tab:ordinary-multiseed-main}a, Qwen reaches
\textbf{36.00$\pm$1.40\%} accuracy, yet every seed solves 6/48 addition
pairs versus 22/48 for its control. For both primary LoRA--Base pairs,
DIS intervals remain below zero under exact-image, scene and union clustering.
NTS intervals include zero under the latter two schemes
\mbox{in Appendix~\ref{sec:cluster-sensitivity}}.

\subsection{Option-Order Robustness}
\label{sec:circular}

Following MirrorQA \citep{liu2026mirrorqa}, we apply four cyclic option
rotations while preserving images and questions. Vanilla scores rotation
zero. Circular requires four correct answers, whereas Consistency requires
four valid answers with identical decoded meanings, correct or incorrect.

Table~\ref{tab:circular-summary}b distinguishes stability from correctness.
GPT leads Circular at \textbf{18.75\%}, whereas Qwen LoRA
is more consistent but reaches \textbf{16.20$\pm$0.53\%} Circular accuracy.
InternVL's higher Vanilla accuracy after adaptation accompanies lower
Circular accuracy and consistency.

\begin{table}[!htb]
  \centering
  \setlength{\abovecaptionskip}{0pt}
  \setlength{\belowcaptionskip}{3pt}
  \caption{\raggedright GPT paired runs (\%, $n=288$). Nearly half of consistent pairs share a wrong answer.}
  \label{tab:same-order-control}
  \begingroup\fontsize{10}{11}\selectfont
  \setlength{\tabcolsep}{4pt}
  \renewcommand{\arraystretch}{1.0}
  \begin{tabular*}{\columnwidth}{@{\extracolsep{\fill}}lrrrr@{}}
    \toprule
    Metric & Repeat & $R_1$ & $R_2$ & $R_3$ \\
    \midrule
    At least one correct & 62.15 & 55.90 & 57.29 & 59.03 \\
    Both correct & 28.47 & 31.25 & 30.56 & 28.12 \\
    Consistency & 53.47 & 59.38 & 54.86 & 51.39 \\
    Consistently wrong & 25.00 & 28.12 & 24.31 & 23.26 \\
    \bottomrule
  \end{tabular*}
  \par\vspace{2pt}
  \parbox{\columnwidth}{\footnotesize\raggedright
  Columns pair $R_0$ with a same-order repeat or rotation.
  Consistency measures agreement in answer meaning.}
  \endgroup
\end{table}

Responses also vary at fixed option order. A repeated GPT run
reaches 45.14\% accuracy versus 45.49\% for rotation zero, but changes
46.53\% of answers. Two-run joint correctness is 28.47\% at fixed order
and 28.12--31.25\% across rotations in Table~\ref{tab:same-order-control}.
Circular scores therefore also reflect execution variability.
Appendix~\ref{sec:gpt-repeat} quantifies uncertainty in these paired comparisons.

\section{Stratified Results and Failure Analysis}
\label{sec:failures}

\begin{table}[!t]
  \centering
  \setlength{\abovecaptionskip}{0pt}
  \setlength{\belowcaptionskip}{5pt}
  \caption{Insufficiency counts out of 48.}
  \label{tab:insufficiency-all-main}
  \begingroup\fontsize{10}{11}\selectfont
  \setlength{\tabcolsep}{3pt}
  \renewcommand{\arraystretch}{1.16}
  \begin{tabular*}{\columnwidth}{@{\extracolsep{\fill}}lrrr@{}}
    \toprule
    \textbf{Model / setting} & \textbf{MSS} & \textbf{Del.} & \textbf{Add.} \\
    \midrule
    Qwen2.5-VL-7B & 2 & 3 & 2 \\
    InternVL3-8B & 13 & 16 & 12 \\
    Qwen3-VL-8B & 7 & 23 & 7 \\
    LLaVA-OneVision-7B & 0 & 1 & 0 \\
    MiniCPM-V 4.5 & 0 & 6 & 0 \\
    Mistral Small 3.2 24B & 29 & 24 & 21 \\
    Qwen2.5-VL-32B & 9 & 18 & 8 \\
    InternVL3.5-14B & 11 & 16 & 9 \\
    Phi-4-Multimodal & 2 & 2 & 2 \\
    Kimi-VL-A3B & 1 & 3 & 2 \\
    \midrule[0.3pt]
    Qwen2.5-VL-7B LoRA & 30 & 38 & 24 \\
    InternVL3-8B LoRA & 22 & 30 & 21 \\
    \midrule[0.3pt]
    GPT-5.6 / 0-shot & 1 & 15 & 2 \\
    \quad 1-shot & 1 & 14 & 2 \\
    \quad 2-shot & 2 & 19 & 1 \\
    \quad 3-shot & 2 & 18 & 1 \\
    \midrule[0.3pt]
    Claude Fable 5.1 / 0-shot & 1 & 10 & 0 \\
    \quad 1-shot & 2 & 5 & 2 \\
    \quad 2-shot & 3 & 15 & 3 \\
    \quad 3-shot & 2 & 15 & 3 \\
    \bottomrule
  \end{tabular*}
  \par\smallskip
  \parbox{\columnwidth}{\footnotesize Unmarked open: Base. Del./Add.: deletion/addition.
  Counts /48 retain invalids: LLaVA addition 10; others 0.}
  \endgroup
\end{table}

\paragraph{Answer changes need not be correct.}
Figure~\ref{fig:typical-errors} illustrates repeated frame answers, spatial
commitment after anchor deletion and answer drift after addition.
Each FQ1 triplet yields three pairs with different gold directions.
Qwen3-VL-8B repeats a direction on 115/144 pairs. GPT three-shot changes
94 pairs but answers both correctly on only 27, distinguishing unchanged
answers from incorrect switches.
From two to three shots, GPT's correct pairs rise from 26 to 27 but complete
triplets fall from five to three. Pairwise gains need not concentrate in
complete groups. Pairwise accuracy can miss failure on the third frame.
Appendix~\ref{sec:fq1-pair-supplement} compares \mbox{all paired outcomes}.

\paragraph{Insufficiency responses extend beyond deletion.}
Both primary LoRA checkpoints increase insufficiency responses across all
conditions in Table~\ref{tab:insufficiency-all-main}. Qwen's insufficiency recall
rises from 6.25\% to 79.17\%, but precision changes from 42.86\% to 41.30\%.
The extra deletion hits therefore come with more false insufficiency
responses on answerable MSS and addition inputs. Joint scoring exposes
this trade-off by requiring success on both sides of the intervention.
Mistral shows the opposite imbalance, selecting insufficiency more often
on MSS than on deletion.
Appendix~\ref{sec:semantic-retention} gives \mbox{the full count breakdown}.

\paragraph{MSS coverage separates intervention outcomes.}
Shared MSS-correct groups account for four of Qwen's five net NTS gains.
InternVL gains one on these groups, offset by changed MSS coverage.
All InternVL joint deletion successes involve removed
anchors in Table~\ref{tab:deletion-reference-main}.

Qwen loses fourteen DIS successes through changed MSS coverage and one
among shared groups. InternVL loses three through each route. Qwen's decline
thus occurs mainly before addition, whereas InternVL also fails to retain
initially correct answers. Similar DIS declines thus have different sources.
Appendix~\ref{sec:shared-mss} provides
\mbox{the paired decomposition}.

\paragraph{Prompt effects vary across executions.}
On twelve triplets excluding removed anchors, two randomized blocks yield
51.39\% agreement with released labels under original wording and 52.78\%
under stable IDs.
Yet NTS falls from 8/24 to 4/24 group--block observations, and 14--15/36
answers change at fixed wording. The small item-level gain thus does not
carry over to matched deletion answers. Appendix~\ref{sec:stable-id-diagnostic}
compares \mbox{each block and condition}.

\section{Conclusion}

FrameEvidence tests spatial reasoning under matched changes in reference
frames and visual evidence. Prompting and adaptation improve item accuracy
without consistent gains in joint success, whereas images most often benefit
MSS--addition pairs. Paired analyses identify incorrect frame switches,
changes in initial MSS coverage and execution instability. Together, these
diagnostics distinguish isolated correct answers from success across matched changes.

\raggedbottom
\section*{Limitations}

FrameEvidence evaluates static, Chinese, four-option spatial reasoning on
96 held-out groups. Scenes can overlap across the image-separated splits,
limiting conclusions about unseen-scene generalization. Metadata cues and image-ordinal changes can
influence responses alongside visual evidence. Production review does not
constitute independent response-level validation of labels or every-member
MSS necessity. NTS therefore assesses agreement with the
released deletion labels.

Model comparisons reflect the tested decoding settings, one demonstration
pool and provider-reported gateway identities. The three-seed deviations
describe training variation. Circular uses its own within-experiment
controls. The human reference is one participant's sampled summary without
individual answers or a documented sample size.

\section*{Ethical Considerations}

The captured and rendered images were supplied by an external data provider
with permission for public release. Language-model
assistance supported editing, scripting and simulated review. The authors
remain responsible for checking the evidence, citations and privacy of the
released materials.

\bibliography{references}
\appendix
\section{Implementation and Scoring Details}
\label{sec:implementation}

\begin{table*}[!t]
\centering
  \setlength{\abovecaptionskip}{0pt}
  \setlength{\belowcaptionskip}{5pt}
  \caption{Model repositories and API aliases used in evaluation.
Repositories identify model families. GPT and Claude names are
provider-reported gateway aliases.}
\label{tab:model-identifiers}

\begingroup\footnotesize
\setlength{\tabcolsep}{3pt}
\begin{tabular*}{\textwidth}{@{\extracolsep{\fill}}ll@{}}
\toprule
Displayed name & Configured repository or gateway identifier \\
\midrule
Qwen2.5-VL-7B & \texttt{Qwen/Qwen2.5-VL-7B-Instruct} \\
InternVL3-8B & \texttt{OpenGVLab/InternVL3-8B-hf} \\
Qwen3-VL-8B & \texttt{Qwen/Qwen3-VL-8B-Instruct} \\
LLaVA-OneVision-7B & \texttt{llava-hf/llava-onevision-qwen2-7b-ov-hf} \\
MiniCPM-V 4.5 & \texttt{openbmb/MiniCPM-V-4\_5} \\
Mistral Small 3.2 24B & \texttt{mistralai/Mistral-Small-3.2-24B-Instruct-2506} \\
Qwen2.5-VL-32B & \texttt{Qwen/Qwen2.5-VL-32B-Instruct} \\
InternVL3.5-14B & \texttt{OpenGVLab/InternVL3\_5-14B-HF} \\
Phi-4-Multimodal & \texttt{microsoft/Phi-4-multimodal-instruct} \\
Kimi-VL-A3B & \texttt{moonshotai/Kimi-VL-A3B-Instruct} \\
\midrule
GPT-5.6 & \texttt{gpt-5.6-sol} \\
Claude Fable 5.1 & \texttt{claude-fable-5-1} \\
\bottomrule
\end{tabular*}
\endgroup

\end{table*}

\paragraph{Evaluation inputs.}
Table~\ref{tab:model-identifiers} lists the evaluated checkpoints and aliases.
Each public item contains ordered image references and a Chinese prompt
asking for exactly one of A, B, C or D. Evaluation identifiers link predictions
to a private mapping with the source sample and group, input mode, gold letter
and option meanings. Group identifiers and gold metadata are not shown to
models. Metrics are computed from the frozen predictions and their matching
input records. All model--mode comparisons use the full question and group
counts.

\paragraph{Offline scoring companion.}
A self-contained Python standard-library scorer reproduces the 20 primary
configurations in Table~\ref{tab:main-results}, eight ordinary multi-seed
configurations in Table~\ref{tab:ordinary-multiseed}, and seven complete
Circular rows in Table~\ref{tab:circular-seeds}. It uses sanitized per-item
records without model calls or image access, preserving all item and group
denominators. Primary, multi-seed and rotation runs remain separate.
It also reproduces GPT's 1,152 Circular responses and 288
same-order repeat responses, including all 16 entries in
Table~\ref{tab:same-order-control}. Running \texttt{python run\_reproduction.py}
from its \texttt{offline\_reproduction} directory writes the recomputed
tables and a verification report. The author-confirmed Qwen Base summary is
reported separately from these recomputed results.

\paragraph{Production review and release correspondence.}
\label{sec:production-qc}
The production database contains 323 PASS and three REWORK actions. Selecting
the latest PASS per group covers all 320 groups and 960 questions. Distinct
annotation and review account assignments are recorded for every group,
with semantic answers visible in the review interface. These are production
approvals rather than independently collected answer sheets. Automatic
comments and checklists do not verify reviewer independence or the
answerability of every single-view deletion.

The task records exactly match the released annotations. Compared
with the latest PASS snapshots, the release preserves all 960 semantic gold
answers and option-meaning sets, and all 1,120 reviewed image files match.
Raw fields nevertheless differ in 917 questions. Of 857 prompt changes,
856 replace a recurring Chinese-character typo and one revises a training
item's anchor (\texttt{fce\_test\_00581}). Option order changes in 540 items
and answer letters in 416, without changing semantic gold. Two anchor fields
also change.
Author adjudication confirms the scored inputs for the four flagged records.

\paragraph{Input and assembly consistency.}
A separate audit compares literal questions, ordered source-image bytes and
options across all 960 items. It finds 959 distinct input signatures. The
only repeated signature belongs to two training items with the same gold
answer. No signature crosses splits or conflicts in its gold letter. All
288 test inputs are distinct. Ignoring option placement alone also yields
no conflicting selected option strings.

All 320 triplets pass the encoded assembly rules. FQ1 preserves ordered
images, options and entity fields across its three frame variants and has
at least two distinct gold meanings. FQ2 preserves question and option
strings, removes or adds exactly one image, and retains the order of the
remaining images. No item repeats identical image content. Assets match
production records. Geometric checks lack camera poses and object coordinates.

\paragraph{Few-shot construction.}
The fixed pool uses training groups \texttt{group\_00152} for FQ1 and
\texttt{group\_00388} for FQ2. FQ1 demonstrations follow camera, object and
scene order. FQ2 demonstrations follow MSS, deletion and addition order.
One-, two- and three-shot conditions take the corresponding prefixes.
The manifest verifies group and exact-image separation from test, although
physical scenes may overlap. All shot counts use this fixed demonstration pool.

\paragraph{Decoding and invalid answers.}
The common output parser extracts an allowed answer letter and records the
raw response, parsing status and API failure separately. An invalid response
is never converted to a guessed answer. In class-wise recall and accuracy
it contributes a false negative for its gold class. Precision uses the
predicted A--D counts, with zero for undefined per-class precision.
Macro-F1 is the mean of the four classwise F1 scores.

A separate full-response parser recovers all 5,750 valid answers among 5,760
main-table records. The ten empty API failures remain incorrect. Reconstructed
answers and point estimates match the frozen results.

\paragraph{Random and human reference rows.}
Uniform random selects one of A--D independently for each question.
The reference row gives exact finite-test expectations. Accuracy and
macro-recall are 25\%, while macro-F1 averages the expected
classwise ratios under the actual supports 75, 70, 69 and 74, yielding
24.93\%. The three-answer and two-answer joint expectations are respectively
$(1/4)^3$ and $(1/4)^2$. Separate binomial calculations and small-test
enumeration verify the scorer.

The human reference is an author-reported P/R/F1/Acc. summary from one
participant answering sampled groups. The retained record contains aggregate
scores without the sample size or item-level responses. We report these
sample-level values and leave joint scores unreported.

\paragraph{Held-out metadata controls.}
\label{sec:heldout-metadata}
\begin{table*}[!t]
\centering
  \setlength{\abovecaptionskip}{0pt}
  \setlength{\belowcaptionskip}{5pt}
  \caption{Train-fitted FQ2 metadata controls (\%). Fixed categorical majority rules use 288 training questions. Dev and test contain 48 and 144 questions, respectively. Joint test metrics use all 48 FQ2 groups. No pixels, question text or condition labels are used.}
\label{tab:heldout-metadata}
\begingroup\footnotesize
\setlength{\tabcolsep}{5pt}
\begin{tabular*}{\textwidth}{@{\extracolsep{\fill}}lrrrrr@{}}
\toprule
Input features & Train Acc. & Dev Acc. & Test Acc. & Test NTS & Test DIS \\
\midrule
Majority letter & 26.39 & 20.83 & 23.61 & 0.00 & 22.92 \\
Image count & 29.51 & 12.50 & 21.53 & 0.00 & 4.17 \\
Count + insufficiency letter & 47.92 & 43.75 & 37.50 & 6.25 & 2.08 \\
\bottomrule\end{tabular*}\endgroup

\end{table*}

The post-hoc controls in Table~\ref{tab:heldout-metadata} fix three categorical
majority rules before computing their held-out scores. They fit only on the
288 FQ2 training questions in 96 groups. Neither the 48 development questions
nor the 144 test questions select rules or features. Ties use alphabetic
letter order, and unseen feature values fall back to the training-global
majority. No held-out feature is unseen. The fixed features are image count
alone and its pairing with the insufficiency option's letter.

The count-plus-schema rule answers 54/144 test questions correctly but solves
only 3/48 MSS/deletion pairs and 1/48 MSS/addition pairs. Conversely, always
predicting the training-majority letter solves 11/48 addition pairs because
the spatial answer letter is shared within each pair. Joint correctness
therefore reveals differences between these rules beyond item accuracy, while the
image-free control shows that shared answer preferences can also yield
joint success. The reproduction materials include fitted rules and predictions.

\paragraph{Adaptation configuration.}
The LoRA checkpoints in Table~\ref{tab:main-results} belong to the
primary comparison. The ordinary multi-seed experiment in
Table~\ref{tab:ordinary-multiseed} evaluates two matched base controls and
six independently trained adapters. Table~\ref{tab:circular-summary} and
Table~\ref{tab:circular-seeds} use those adapted checkpoints in separate
rotation runs. All three experiments use the same 288 test questions but
separate predictions. Vanilla uses Circular's own rotation-zero predictions.

\begin{table*}[!t]
  \centering
  \setlength{\abovecaptionskip}{0pt}
  \setlength{\belowcaptionskip}{5pt}
  \caption{Ordinary all-images multi-seed results (\%). GEM denotes FQ1-GEM. Classification uses 288 questions and joint criteria use 48 groups per family. Means and sample deviations use three LoRA seeds. These ordinary runs are separate from Circular rotation zero.}
  \label{tab:ordinary-multiseed}

  \begingroup\footnotesize
  \setlength{\tabcolsep}{2pt}
  \begin{tabular*}{\textwidth}{@{\extracolsep{\fill}}llrrrrrrr@{}}
    \toprule
    Family & Run & P & R & F1 & Acc. & GEM & NTS & DIS \\
    \midrule
    InternVL3-8B & Base & 30.34 & 30.54 & 29.36 & 30.21 & 2.08 & 8.33 & 20.83 \\
     & 20260722 & 34.32 & 33.20 & 31.50 & 32.99 & 0.00 & 10.42 & 14.58 \\
     & 20260723 & 32.09 & 31.19 & 29.11 & 30.90 & 0.00 & 8.33 & 10.42 \\
     & 20260724 & 33.85 & 33.64 & 31.85 & 33.33 & 2.08 & 10.42 & 12.50 \\
     & Mean & $33.42{\pm}1.17$ & $32.68{\pm}1.31$ & $30.82{\pm}1.49$ & $32.41{\pm}1.31$ & $0.69{\pm}1.20$ & $9.72{\pm}1.20$ & $12.50{\pm}2.08$ \\
    \midrule
    Qwen2.5-VL-7B & Base & 30.87 & 29.77 & 26.35 & 29.17 & 0.00 & 2.08 & 45.83 \\
     & 20260722 & 35.74 & 35.88 & 35.00 & 35.76 & 0.00 & 8.33 & 12.50 \\
     & 20260723 & 35.10 & 34.85 & 33.99 & 34.72 & 0.00 & 8.33 & 12.50 \\
     & 20260724 & 37.53 & 37.57 & 37.10 & 37.50 & 0.00 & 14.58 & 12.50 \\
     & Mean & $36.12{\pm}1.26$ & $36.10{\pm}1.37$ & $35.36{\pm}1.59$ & $36.00{\pm}1.40$ & $0.00{\pm}0.00$ & $10.42{\pm}3.61$ & $12.50{\pm}0.00$ \\
    \bottomrule
  \end{tabular*}
  \endgroup
  
\end{table*}

Training uses LoRA rank 32, alpha 16 and zero
dropout, with the vision tower and multimodal projector frozen. It specifies
a 4,096-token cutoff, learning rate $2\times10^{-5}$, six epochs, cosine
schedule, warmup ratio 0.05, and batch size one per device with eight
accumulation steps on two devices. This gives effective batch size 16.
The training export contains 576 train and 96 development items and excludes
test responses. Seeds 20260722--20260724 are used in the three-seed experiment.
All six runs complete 216 optimization steps over six epochs. Development
loss selects step 180 for Qwen seeds 20260722 and 20260723, and step 216
for the remaining runs. Training and development exports contain no test
sample identifiers.
The companion reproduces scores from saved predictions and records the
training configuration. Circular records contain four option rotations per
source item.
Their prompts and image order match the ordinary inputs at rotation zero.
Separate executions nevertheless disagree on some items, so Vanilla uses
the rotation run's own predictions.

\paragraph{Circular metrics.}
Let $a_{ir}$ be the prediction for source item $i$ at rotation $r$, mapped
back to the original option identity, and let $y_i$ be its gold identity.
For each item, let $C_i=\prod_{r=0}^{3}\mathbb{1}[a_{ir}=y_i]$ indicate
correctness across all four rotations.
Let $S_i=\mathbb{1}[a_{i0}=a_{i1}=a_{i2}=a_{i3}]$ indicate consistency.
With $N=288$, we compute
\begin{equation}
\mathrm{Circular}=\frac{\sum_i C_i}{N},\quad
\mathrm{Cons.}=\frac{\sum_i S_i}{N}.
\end{equation}
All four predictions must be valid for consistency to receive credit.
The reference implementation additionally reports each rotation and the
mean over rotations. Table~\ref{tab:circular-seeds} retains the per-seed
scores. Family means and sample deviations are calculated from unrounded
item counts. All six adapted checkpoints and the InternVL Base control have
complete retained predictions. Qwen Base is reported from the final run
summary rather than recomputed from the retained prediction copy.
The GPT supplement adds 1,152 valid zero-shot responses over the same 288
source questions. Its four rotations yield 131, 120, 122 and 120 correct
answers. There are 54 fully correct sets and 78 semantically consistent sets.
Its Vanilla count is therefore 131/288, versus 117/288 in the primary
all-images run.

\begin{table*}[!t]
  \centering
  \setlength{\abovecaptionskip}{0pt}
  \setlength{\belowcaptionskip}{5pt}
  \caption{Per-seed open-weight and zero-shot GPT Circular scores (\%) for 288 items and four cyclic rotations.
Drop uses percentage points. Qwen Base is summary-reported. Other rows
use 1,152 retained valid responses each.}
\label{tab:circular-seeds}

  \begingroup\footnotesize
  \setlength{\tabcolsep}{4pt}
  \begin{tabular*}{\textwidth}{@{\extracolsep{\fill}}llrrrrr@{}}
    \toprule
    Family & Seed / control & Vanilla & Mean rot. & Circular & Consistency & Drop \\
    \midrule
    InternVL3-8B & Base & 30.21 & 28.04 & 7.64 & 23.61 & 22.57 \\
     & 20260722 & 33.33 & 31.42 & 7.29 & 19.10 & 26.04 \\
     & 20260723 & 33.68 & 31.08 & 5.56 & 17.71 & 28.12 \\
     & 20260724 & 31.25 & 30.30 & 5.21 & 15.97 & 26.04 \\
    Qwen2.5-VL-7B & Base & 14.93 & 18.75 & 2.08 & 5.90 & 12.85 \\
     & 20260722 & 34.38 & 35.76 & 16.67 & 33.68 & 17.71 \\
     & 20260723 & 34.72 & 35.42 & 15.62 & 32.99 & 19.10 \\
     & 20260724 & 36.46 & 36.81 & 16.32 & 31.25 & 20.14 \\
    \midrule
    GPT-5.6 & 0-shot & 45.49 & 42.80 & 18.75 & 27.08 & 26.74 \\
    \bottomrule
  \end{tabular*}
  \endgroup

\end{table*}

Qwen Base's rounded summary also matches scores from the incomplete copy
when its 446 failed calls count as incorrect. That copy cannot independently
confirm completion. We retain the author-confirmed summary as a reference
in Tables~\ref{tab:circular-summary} and~\ref{tab:circular-seeds}. It is separate
from the independently recomputed paired estimates.

\paragraph{GPT cross-run stability.}
Pairing ordinary all-images and Circular rotation zero on all 288 source
items gives 79 correct in both executions, 38 correct only in ordinary,
52 correct only in rotation zero, and 119 incorrect in both. Their answer
letters differ on 136 items (47.22\%). Prepared prompts, option order,
ordered image references and retained decoding configurations match.
Archived wire requests and backend revisions are unavailable, so this is
a descriptive cross-run comparison, not an isolated decoding experiment.
We resample whole clusters with replacement 20,000 times, using seed
20261010 and item-weighted ratios. Across the 13 union components defined
in Appendix~\ref{sec:cluster-sensitivity}, the 95\% percentile interval for
disagreement is [41.33, 56.61]\%. The accuracy difference is 4.86 points
with interval [$-3.67$, 16.67]. These intervals quantify sensitivity to item
composition under this clustering, not repeated API execution.

\paragraph{Same-order control.}
\label{sec:gpt-repeat}
\begin{table*}[!t]
  \centering
  \caption{Paired GPT same-order control with 95\% bootstrap intervals.
  Rates are percentages and differences are percentage points.
  Rotation contrasts subtract the corresponding $R_0$--$R_j$ pair
  from the $R_0$--repeat pair.}
  \label{tab:repeat-uncertainty}
  \small
  \setlength{\tabcolsep}{5pt}
  \begin{tabular*}{\textwidth}{@{\extracolsep{\fill}}lrrr@{}}
    \toprule
    Quantity & Estimate & 96 triplets & 13 union components \\
    \midrule
    Repeat accuracy & 45.14 & [39.58, 50.69] & [39.09, 50.56] \\
    Primary--repeat disagreement & 44.44 & [38.54, 50.69] & [37.92, 54.49] \\
    $R_0$--repeat disagreement & 46.53 & [40.28, 52.78] & [43.18, 51.96] \\
    Repeat minus primary accuracy & 4.51 & [$-1.74$, 10.76] & [$-1.99$, 14.29] \\
    Repeat minus $R_0$ accuracy & $-0.35$ & [$-6.94$, 6.25] & [$-12.28$, 6.98] \\
    Same-order pair both correct & 28.47 & [23.26, 33.68] & [23.50, 33.90] \\
    Same-order pair consistency & 53.47 & [47.22, 59.72] & [48.04, 56.82] \\
    \midrule
    Both correct, difference from $R_0$--$R_1$ & $-2.78$ & [$-7.29$, 1.39] & [$-10.12$, 1.55] \\
    Both correct, difference from $R_0$--$R_2$ & $-2.08$ & [$-6.60$, 2.43] & [$-8.84$, 3.03] \\
    Both correct, difference from $R_0$--$R_3$ & 0.35 & [$-4.17$, 4.86] & [$-7.21$, 5.51] \\
    Consistency, difference from $R_0$--$R_1$ & $-5.90$ & [$-13.19$, 1.04] & [$-15.33$, $-0.65$] \\
    Consistency, difference from $R_0$--$R_2$ & $-1.39$ & [$-8.33$, 5.56] & [$-9.60$, 7.34] \\
    Consistency, difference from $R_0$--$R_3$ & 2.08 & [$-4.51$, 8.33] & [$-7.04$, 7.68] \\
    \bottomrule
  \end{tabular*}
\end{table*}

We repeat the same 288 zero-shot all-images queries with identical
GPT request settings. All responses are valid. The repeat yields
130 correct answers, with P/R/F1/Acc of 45.66/45.31/45.06/45.14\%.
Its FQ1-GEM, NTS and DIS are 4.17\%, 20.83\% and 31.25\%.
Paired with the primary run, 77
answers are correct in both, 40 only in the primary, 53 only in the repeat,
and 118 in neither. Predictions change on 128/288 items. Pairing with
rotation zero gives counts of 82, 49, 48 and 109, with 134 changed answers.

Table~\ref{tab:same-order-control} compares two-answer sets sharing the
same rotation-zero output. The same-order pair is jointly correct on 82
items and semantically consistent on 154. Pairing rotation zero with
rotations one, two and three gives joint counts of 90, 88 and 81 and
consistency counts of 171, 158 and 148. Holding set size fixed removes
the two-versus-four conjunction difference when assessing stability.

We resample paired observations in 20,000 whole-cluster bootstrap draws
with seed 20261010, using both 96 triplets and 13 image/scene union
components. Table~\ref{tab:repeat-uncertainty} reports estimates with both
95\% intervals. Every joint-correctness contrast with a rotation pair
includes zero under both schemes. The consistency contrast with rotation
one is below zero under union clustering but includes zero under triplet
clustering. These intervals describe question-composition sensitivity.
The repeat and rotation runs were executed at different times,
so temporal service changes and decoding variation remain combined.

\section{Release Composition}
\label{sec:appendix-composition}

\begin{figure*}[!t]
  \centering
  \includegraphics[width=\textwidth]{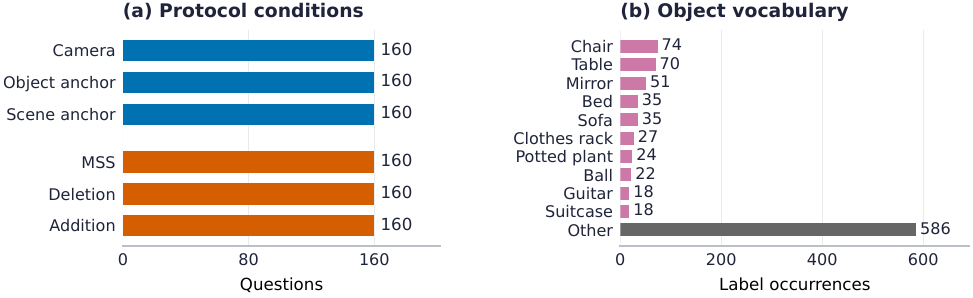}
  \caption{Release composition. Each of six conditions contains 160 questions.
  Object-label occurrences are counted once per matched group across
  126 distinct labels. Other pools labels outside the top ten.}
  \label{fig:dataset-composition}
\end{figure*}

Figure~\ref{fig:dataset-composition} shows condition balance and source
object-label frequencies, counted once per group. The Other category
aggregates labels outside the ten most frequent types.

\section{Complete Input Controls}
\label{sec:input-controls}

Table~\ref{tab:input-ablation} includes twelve open-weight configurations and
GPT zero-shot. Original gold labels remain fixed even when an ablation
removes information needed to infer them. GPT retains 288 valid responses
per mode and uses the primary table's all-images predictions. Its companion
provides sanitized per-item inputs, predictions and exact counts.
\begin{table*}[!t]
\centering
  \setlength{\abovecaptionskip}{0pt}
  \setlength{\belowcaptionskip}{5pt}
  \caption{Visual-input ablations on 288 test questions (\%). Questions, options and gold answers are held fixed across all-images, first-image and text-only modes.}
\label{tab:input-ablation}

\begingroup\footnotesize
\setlength{\tabcolsep}{1.4pt}
\begin{tabular*}{\textwidth}{@{\extracolsep{\fill}}lrrrrrrrrrrrr@{}}
\toprule
Model & \multicolumn{4}{c}{All images} & \multicolumn{4}{c}{First image} & \multicolumn{4}{c}{Text only} \\
\cmidrule(lr){2-5}\cmidrule(lr){6-9}\cmidrule(l){10-13}
 & Acc. & GEM & NTS & DIS & Acc. & GEM & NTS & DIS & Acc. & GEM & NTS & DIS \\
\midrule
Qwen2.5-VL-7B Base & 28.82 & 0.00 & 2.08 & 43.75 & 29.17 & 0.00 & 2.08 & 37.50 & 31.94 & 0.00 & 0.00 & 0.00 \\
Qwen2.5-VL-7B LoRA & 35.42 & 0.00 & 12.50 & 12.50 & 31.25 & 0.00 & 0.00 & 0.00 & 32.29 & 2.08 & 0.00 & 0.00 \\
InternVL3-8B Base & 29.51 & 2.08 & 8.33 & 20.83 & 27.78 & 0.00 & 6.25 & 29.17 & 30.90 & 2.08 & 0.00 & 8.33 \\
InternVL3-8B LoRA & 32.64 & 0.00 & 8.33 & 8.33 & 27.78 & 0.00 & 4.17 & 10.42 & 30.21 & 0.00 & 0.00 & 0.00 \\
Qwen3-VL-8B Base & 34.03 & 0.00 & 14.58 & 39.58 & 28.82 & 0.00 & 8.33 & 10.42 & 29.51 & 0.00 & 0.00 & 6.25 \\
LLaVA-OneVision-7B Base & 29.17 & 0.00 & 0.00 & 37.50 & 31.60 & 0.00 & 0.00 & 50.00 & 29.17 & 0.00 & 0.00 & 43.75 \\
MiniCPM-V 4.5 Base & 32.99 & 0.00 & 4.17 & 47.92 & 31.94 & 0.00 & 8.33 & 43.75 & 29.17 & 0.00 & 0.00 & 33.33 \\
Mistral Small 3.2 24B Base & 25.69 & 0.00 & 2.08 & 8.33 & 29.51 & 0.00 & 6.25 & 31.25 & 29.51 & 0.00 & 0.00 & 0.00 \\
Qwen2.5-VL-32B Base & 33.33 & 0.00 & 6.25 & 35.42 & 26.74 & 0.00 & 8.33 & 16.67 & 30.90 & 0.00 & 0.00 & 0.00 \\
InternVL3.5-14B Base & 28.12 & 0.00 & 6.25 & 25.00 & 24.65 & 0.00 & 8.33 & 8.33 & 30.21 & 0.00 & 0.00 & 14.58 \\
Phi-4-Multimodal Base & 26.39 & 0.00 & 0.00 & 35.42 & 28.12 & 0.00 & 4.17 & 37.50 & 28.82 & 2.08 & 0.00 & 18.75 \\
Kimi-VL-A3B Base & 29.17 & 0.00 & 0.00 & 39.58 & 30.21 & 2.08 & 6.25 & 31.25 & 25.00 & 0.00 & 0.00 & 35.42 \\
\midrule
GPT-5.6 / 0-shot & 40.62 & 4.17 & 14.58 & 35.42 & 29.51 & 0.00 & 12.50 & 8.33 & 31.94 & 0.00 & 0.00 & 0.00 \\
\bottomrule
\end{tabular*}
\endgroup

\end{table*}

\section{Stratified Results and Paired Contrasts}
\label{sec:stratified-results}

Tables~\ref{tab:frameevidence-count-strata},
\ref{tab:frameevidence-relation-strata} and~\ref{tab:frameevidence-scene-strata}
break down all-images results by image count, relation and scene.
Each stratum contains different instances, so the contrasts are descriptive.
Table~\ref{tab:frameevidence-fq2-frame-strata} uses FQ2 annotation tags,
not independently validated frame categories. A lexical audit finds camera-heading
references in 30/48 MSS prompts with non-camera tags, covering 14 of 16
object-tagged and all 16 world-tagged groups.

\subsection{Geometry-associated differences}
Table~\ref{tab:geometry-summary} shows that axial relations favor DIS across
most configurations, but NTS has no median gap. Granularity is associated
more consistently with addition than deletion-label success.
Shared scene-dependent groups preclude treating configurations as
independent replications.
\ifdefined\FEpairedtables\else\begin{table}[tb]\fi
\centering
  \setlength{\abovecaptionskip}{0pt}
  \setlength{\belowcaptionskip}{5pt}
  \caption{Axial--quadrant gaps over 20 configurations (24 groups per stratum). Signs count positive, tied and negative gaps. Median $\Delta$ is in percentage points.}
\label{tab:geometry-summary}

\begingroup\footnotesize
\setlength{\tabcolsep}{3pt}
\begin{tabular*}{\columnwidth}{@{\extracolsep{\fill}}lrrrr@{}}
\toprule
Metric & $>$ & $=$ & $<$ & Median $\Delta$ \\
\midrule
FQ1-GEM & 2 & 13 & 5 & 0.0 \\
NTS & 5 & 8 & 7 & 0.0 \\
DIS & 17 & 3 & 0 & +20.8 \\
\bottomrule
\end{tabular*}
\endgroup

\ifdefined\FEpairedtables\else\end{table}\fi

\begin{table*}[!t]
\centering
  \setlength{\abovecaptionskip}{0pt}
  \setlength{\belowcaptionskip}{5pt}
  \caption{Evidence-set size strata (\%). Each column gives its group denominator $n$.}
\label{tab:frameevidence-count-strata}

\begingroup\footnotesize
\setlength{\tabcolsep}{2.5pt}
\begin{tabular*}{\textwidth}{@{\extracolsep{\fill}}lrrrrrrrrr@{}}
\toprule
Model & \multicolumn{3}{c}{FQ1-GEM} & \multicolumn{3}{c}{NTS} & \multicolumn{3}{c}{DIS} \\
\cmidrule(lr){2-4}\cmidrule(lr){5-7}\cmidrule(lr){8-10}
 & \shortstack{2\\$n=15$} & \shortstack{3\\$n=16$} & \shortstack{4\\$n=17$} & \shortstack{2\\$n=17$} & \shortstack{3\\$n=16$} & \shortstack{4\\$n=15$} & \shortstack{2\\$n=17$} & \shortstack{3\\$n=16$} & \shortstack{4\\$n=15$} \\
\midrule
Claude Fable 5.1 (0-shot) & 0.0 & 0.0 & 0.0 & 11.8 & 12.5 & 0.0 & 23.5 & 56.2 & 46.7 \\
Claude Fable 5.1 (1-shot) & 6.7 & 12.5 & 0.0 & 0.0 & 6.2 & 0.0 & 29.4 & 37.5 & 46.7 \\
Claude Fable 5.1 (2-shot) & 0.0 & 12.5 & 5.9 & 17.6 & 18.8 & 0.0 & 35.3 & 37.5 & 40.0 \\
Claude Fable 5.1 (3-shot) & 0.0 & 6.2 & 0.0 & 11.8 & 12.5 & 0.0 & 29.4 & 56.2 & 33.3 \\
GPT-5.6 (0-shot) & 6.7 & 0.0 & 5.9 & 17.6 & 25.0 & 0.0 & 35.3 & 25.0 & 46.7 \\
GPT-5.6 (1-shot) & 0.0 & 0.0 & 17.6 & 11.8 & 31.2 & 6.7 & 41.2 & 43.8 & 40.0 \\
GPT-5.6 (2-shot) & 0.0 & 12.5 & 17.6 & 23.5 & 43.8 & 0.0 & 41.2 & 56.2 & 33.3 \\
GPT-5.6 (3-shot) & 6.7 & 12.5 & 0.0 & 29.4 & 25.0 & 6.7 & 41.2 & 50.0 & 53.3 \\
InternVL3-8B Base & 6.7 & 0.0 & 0.0 & 5.9 & 6.2 & 13.3 & 17.6 & 25.0 & 20.0 \\
InternVL3-8B LoRA & 0.0 & 0.0 & 0.0 & 11.8 & 6.2 & 6.7 & 11.8 & 12.5 & 0.0 \\
InternVL3.5-14B & 0.0 & 0.0 & 0.0 & 5.9 & 6.2 & 6.7 & 17.6 & 18.8 & 40.0 \\
Kimi-VL-A3B & 0.0 & 0.0 & 0.0 & 0.0 & 0.0 & 0.0 & 41.2 & 43.8 & 33.3 \\
LLaVA-OneVision-7B & 0.0 & 0.0 & 0.0 & 0.0 & 0.0 & 0.0 & 41.2 & 43.8 & 26.7 \\
MiniCPM-V 4.5 & 0.0 & 0.0 & 0.0 & 5.9 & 6.2 & 0.0 & 41.2 & 37.5 & 66.7 \\
Mistral Small 3.2 24B & 0.0 & 0.0 & 0.0 & 5.9 & 0.0 & 0.0 & 11.8 & 0.0 & 13.3 \\
Phi-4-Multimodal & 0.0 & 0.0 & 0.0 & 0.0 & 0.0 & 0.0 & 29.4 & 31.2 & 46.7 \\
Qwen2.5-VL-32B & 0.0 & 0.0 & 0.0 & 5.9 & 12.5 & 0.0 & 35.3 & 31.2 & 40.0 \\
Qwen2.5-VL-7B Base & 0.0 & 0.0 & 0.0 & 5.9 & 0.0 & 0.0 & 47.1 & 56.2 & 26.7 \\
Qwen2.5-VL-7B LoRA & 0.0 & 0.0 & 0.0 & 0.0 & 25.0 & 13.3 & 0.0 & 25.0 & 13.3 \\
Qwen3-VL-8B & 0.0 & 0.0 & 0.0 & 23.5 & 12.5 & 6.7 & 29.4 & 37.5 & 53.3 \\
\bottomrule
\end{tabular*}
\par\smallskip\parbox{\textwidth}{\footnotesize All configurations use the all-images test, with strata comparing different held-out groups.}
\endgroup

\end{table*}

\begin{table*}[!t]
\centering
  \setlength{\abovecaptionskip}{0pt}
  \setlength{\belowcaptionskip}{5pt}
  \caption{Axial and quadrant relation strata (\%). Each column gives its group denominator $n$.}
\label{tab:frameevidence-relation-strata}

\begingroup\footnotesize
\setlength{\tabcolsep}{2.5pt}
\begin{tabular*}{\textwidth}{@{\extracolsep{\fill}}lrrrrrr@{}}
\toprule
Model & \multicolumn{2}{c}{FQ1-GEM} & \multicolumn{2}{c}{NTS} & \multicolumn{2}{c}{DIS} \\
\cmidrule(lr){2-3}\cmidrule(lr){4-5}\cmidrule(lr){6-7}
 & \shortstack{Axial\\$n=24$} & \shortstack{Quadrant\\$n=24$} & \shortstack{Axial\\$n=24$} & \shortstack{Quadrant\\$n=24$} & \shortstack{Axial\\$n=24$} & \shortstack{Quadrant\\$n=24$} \\
\midrule
Claude Fable 5.1 (0-shot) & 0.0 & 0.0 & 16.7 & 0.0 & 54.2 & 29.2 \\
Claude Fable 5.1 (1-shot) & 4.2 & 8.3 & 0.0 & 4.2 & 45.8 & 29.2 \\
Claude Fable 5.1 (2-shot) & 4.2 & 8.3 & 12.5 & 12.5 & 54.2 & 20.8 \\
Claude Fable 5.1 (3-shot) & 0.0 & 4.2 & 8.3 & 8.3 & 50.0 & 29.2 \\
GPT-5.6 (0-shot) & 4.2 & 4.2 & 16.7 & 12.5 & 45.8 & 25.0 \\
GPT-5.6 (1-shot) & 8.3 & 4.2 & 8.3 & 25.0 & 41.7 & 41.7 \\
GPT-5.6 (2-shot) & 8.3 & 12.5 & 33.3 & 12.5 & 50.0 & 37.5 \\
GPT-5.6 (3-shot) & 4.2 & 8.3 & 25.0 & 16.7 & 58.3 & 37.5 \\
InternVL3-8B Base & 4.2 & 0.0 & 8.3 & 8.3 & 29.2 & 12.5 \\
InternVL3-8B LoRA & 0.0 & 0.0 & 8.3 & 8.3 & 12.5 & 4.2 \\
InternVL3.5-14B & 0.0 & 0.0 & 8.3 & 4.2 & 37.5 & 12.5 \\
Kimi-VL-A3B & 0.0 & 0.0 & 0.0 & 0.0 & 66.7 & 12.5 \\
LLaVA-OneVision-7B & 0.0 & 0.0 & 0.0 & 0.0 & 50.0 & 25.0 \\
MiniCPM-V 4.5 & 0.0 & 0.0 & 0.0 & 8.3 & 58.3 & 37.5 \\
Mistral Small 3.2 24B & 0.0 & 0.0 & 0.0 & 4.2 & 8.3 & 8.3 \\
Phi-4-Multimodal & 0.0 & 0.0 & 0.0 & 0.0 & 54.2 & 16.7 \\
Qwen2.5-VL-32B & 0.0 & 0.0 & 4.2 & 8.3 & 50.0 & 20.8 \\
Qwen2.5-VL-7B Base & 0.0 & 0.0 & 0.0 & 4.2 & 50.0 & 37.5 \\
Qwen2.5-VL-7B LoRA & 0.0 & 0.0 & 12.5 & 12.5 & 12.5 & 12.5 \\
Qwen3-VL-8B & 0.0 & 0.0 & 12.5 & 16.7 & 50.0 & 29.2 \\
\bottomrule
\end{tabular*}
\par\smallskip\parbox{\textwidth}{\footnotesize All configurations use the all-images test, with strata comparing different held-out groups.}
\endgroup

\end{table*}

\begin{table*}[!t]
\centering
  \setlength{\abovecaptionskip}{0pt}
  \setlength{\belowcaptionskip}{5pt}
  \caption{Scene-category strata (\%). Each column gives its group denominator $n$.}
\label{tab:frameevidence-scene-strata}

\begingroup\footnotesize
\setlength{\tabcolsep}{2.5pt}
\begin{tabular*}{\textwidth}{@{\extracolsep{\fill}}lrrrrrrrrr@{}}
\toprule
Model & \multicolumn{3}{c}{FQ1-GEM} & \multicolumn{3}{c}{NTS} & \multicolumn{3}{c}{DIS} \\
\cmidrule(lr){2-4}\cmidrule(lr){5-7}\cmidrule(lr){8-10}
 & \shortstack{Large\\$n=15$} & \shortstack{Tabletop\\$n=17$} & \shortstack{Outdoor\\$n=16$} & \shortstack{Large\\$n=16$} & \shortstack{Tabletop\\$n=17$} & \shortstack{Outdoor\\$n=15$} & \shortstack{Large\\$n=16$} & \shortstack{Tabletop\\$n=17$} & \shortstack{Outdoor\\$n=15$} \\
\midrule
Claude Fable 5.1 (0-shot) & 0.0 & 0.0 & 0.0 & 12.5 & 5.9 & 6.7 & 50.0 & 23.5 & 53.3 \\
Claude Fable 5.1 (1-shot) & 13.3 & 5.9 & 0.0 & 6.2 & 0.0 & 0.0 & 31.2 & 41.2 & 40.0 \\
Claude Fable 5.1 (2-shot) & 6.7 & 5.9 & 6.2 & 12.5 & 11.8 & 13.3 & 25.0 & 35.3 & 53.3 \\
Claude Fable 5.1 (3-shot) & 0.0 & 5.9 & 0.0 & 18.8 & 5.9 & 0.0 & 37.5 & 35.3 & 46.7 \\
GPT-5.6 (0-shot) & 0.0 & 11.8 & 0.0 & 18.8 & 17.6 & 6.7 & 37.5 & 41.2 & 26.7 \\
GPT-5.6 (1-shot) & 6.7 & 5.9 & 6.2 & 31.2 & 0.0 & 20.0 & 56.2 & 29.4 & 40.0 \\
GPT-5.6 (2-shot) & 13.3 & 17.6 & 0.0 & 50.0 & 5.9 & 13.3 & 75.0 & 17.6 & 40.0 \\
GPT-5.6 (3-shot) & 0.0 & 5.9 & 12.5 & 31.2 & 17.6 & 13.3 & 56.2 & 47.1 & 40.0 \\
InternVL3-8B Base & 0.0 & 5.9 & 0.0 & 6.2 & 5.9 & 13.3 & 37.5 & 11.8 & 13.3 \\
InternVL3-8B LoRA & 0.0 & 0.0 & 0.0 & 12.5 & 11.8 & 0.0 & 6.2 & 11.8 & 6.7 \\
InternVL3.5-14B & 0.0 & 0.0 & 0.0 & 6.2 & 5.9 & 6.7 & 25.0 & 23.5 & 26.7 \\
Kimi-VL-A3B & 0.0 & 0.0 & 0.0 & 0.0 & 0.0 & 0.0 & 31.2 & 29.4 & 60.0 \\
LLaVA-OneVision-7B & 0.0 & 0.0 & 0.0 & 0.0 & 0.0 & 0.0 & 25.0 & 29.4 & 60.0 \\
MiniCPM-V 4.5 & 0.0 & 0.0 & 0.0 & 6.2 & 5.9 & 0.0 & 43.8 & 41.2 & 60.0 \\
Mistral Small 3.2 24B & 0.0 & 0.0 & 0.0 & 0.0 & 0.0 & 6.7 & 0.0 & 17.6 & 6.7 \\
Phi-4-Multimodal & 0.0 & 0.0 & 0.0 & 0.0 & 0.0 & 0.0 & 31.2 & 29.4 & 46.7 \\
Qwen2.5-VL-32B & 0.0 & 0.0 & 0.0 & 6.2 & 11.8 & 0.0 & 25.0 & 47.1 & 33.3 \\
Qwen2.5-VL-7B Base & 0.0 & 0.0 & 0.0 & 0.0 & 0.0 & 6.7 & 31.2 & 41.2 & 60.0 \\
Qwen2.5-VL-7B LoRA & 0.0 & 0.0 & 0.0 & 12.5 & 11.8 & 13.3 & 25.0 & 5.9 & 6.7 \\
Qwen3-VL-8B & 0.0 & 0.0 & 0.0 & 18.8 & 11.8 & 13.3 & 37.5 & 41.2 & 40.0 \\
\bottomrule
\end{tabular*}
\par\smallskip\parbox{\textwidth}{\footnotesize All configurations use the all-images test, with strata comparing different held-out groups.}
\endgroup

\end{table*}

\begin{table*}[!t]
\centering
  \setlength{\abovecaptionskip}{0pt}
  \setlength{\belowcaptionskip}{5pt}
  \caption{FQ2 annotation-tag strata (\%). Each column gives its group denominator $n$.}
\label{tab:frameevidence-fq2-frame-strata}

\begingroup\footnotesize
\setlength{\tabcolsep}{2.5pt}
\begin{tabular*}{\textwidth}{@{\extracolsep{\fill}}lrrrrrr@{}}
\toprule
Model & \multicolumn{3}{c}{NTS} & \multicolumn{3}{c}{DIS} \\
\cmidrule(lr){2-4}\cmidrule(lr){5-7}
 & \shortstack{Camera tag\\$n=16$} & \shortstack{Object tag\\$n=16$} & \shortstack{World tag\\$n=16$} & \shortstack{Camera tag\\$n=16$} & \shortstack{Object tag\\$n=16$} & \shortstack{World tag\\$n=16$} \\
\midrule
Claude Fable 5.1 (0-shot) & 12.5 & 6.2 & 6.2 & 43.8 & 37.5 & 43.8 \\
Claude Fable 5.1 (1-shot) & 6.2 & 0.0 & 0.0 & 43.8 & 31.2 & 37.5 \\
Claude Fable 5.1 (2-shot) & 18.8 & 6.2 & 12.5 & 50.0 & 31.2 & 31.2 \\
Claude Fable 5.1 (3-shot) & 6.2 & 12.5 & 6.2 & 43.8 & 43.8 & 31.2 \\
GPT-5.6 (0-shot) & 12.5 & 12.5 & 18.8 & 37.5 & 31.2 & 37.5 \\
GPT-5.6 (1-shot) & 12.5 & 18.8 & 18.8 & 50.0 & 37.5 & 37.5 \\
GPT-5.6 (2-shot) & 25.0 & 12.5 & 31.2 & 50.0 & 37.5 & 43.8 \\
GPT-5.6 (3-shot) & 37.5 & 12.5 & 12.5 & 50.0 & 37.5 & 56.2 \\
InternVL3-8B Base & 6.2 & 12.5 & 6.2 & 12.5 & 31.2 & 18.8 \\
InternVL3-8B LoRA & 6.2 & 12.5 & 6.2 & 6.2 & 12.5 & 6.2 \\
InternVL3.5-14B & 12.5 & 6.2 & 0.0 & 31.2 & 37.5 & 6.2 \\
Kimi-VL-A3B & 0.0 & 0.0 & 0.0 & 31.2 & 43.8 & 43.8 \\
LLaVA-OneVision-7B & 0.0 & 0.0 & 0.0 & 50.0 & 25.0 & 37.5 \\
MiniCPM-V 4.5 & 0.0 & 12.5 & 0.0 & 43.8 & 50.0 & 50.0 \\
Mistral Small 3.2 24B & 6.2 & 0.0 & 0.0 & 12.5 & 12.5 & 0.0 \\
Phi-4-Multimodal & 0.0 & 0.0 & 0.0 & 37.5 & 31.2 & 37.5 \\
Qwen2.5-VL-32B & 0.0 & 12.5 & 6.2 & 37.5 & 43.8 & 25.0 \\
Qwen2.5-VL-7B Base & 0.0 & 0.0 & 6.2 & 31.2 & 43.8 & 56.2 \\
Qwen2.5-VL-7B LoRA & 0.0 & 18.8 & 18.8 & 0.0 & 18.8 & 18.8 \\
Qwen3-VL-8B & 6.2 & 31.2 & 6.2 & 25.0 & 68.8 & 25.0 \\
\bottomrule
\end{tabular*}
\par\smallskip\parbox{\textwidth}{\footnotesize All-images test grouped by annotation tags. Thirty of 48 MSS prompts name a camera heading despite a non-camera tag. These tags have not been independently validated as frame categories.}
\endgroup

\end{table*}

\subsection{Unconditional response distributions}
Table~\ref{tab:insufficiency-all-main} counts insufficient-information
responses in every condition, with all 48 groups retained. Option meanings,
not letters, define this category. All 576 FQ2 predictions from the four
primary Base/LoRA checkpoints are valid.
For Qwen, the Base-to-LoRA spatial-to-insufficiency switches number 28, 35
and 22 for MSS, deletion and addition, with no reverse switches. InternVL
has 10, 14 and nine such switches, with one reverse switch in MSS and none
elsewhere. The paired answers reveal a broader shift toward insufficiency
across all three conditions.

\paragraph{Conditional failures.}
Table~\ref{tab:frameevidence-failure-analysis} conditions deletion and
addition outcomes on an initially correct MSS answer, so denominators
differ by model. Unconditional NTS and DIS always use all 48 FQ2 groups.
Together, these views separate errors on the initial question from failures
after the intervention.

\begin{table*}[!t]
\centering
  \setlength{\abovecaptionskip}{0pt}
  \setlength{\belowcaptionskip}{5pt}
  \caption{Failure profiles on the all-images test (\%). Frame invariance is unchanged semantic output despite different gold directions.}
\label{tab:frameevidence-failure-analysis}

\begingroup\footnotesize
\setlength{\tabcolsep}{2.7pt}
\begin{tabular*}{\textwidth}{@{\extracolsep{\fill}}lrrrrrrrr@{}}
\toprule
Model & Frame & $m$ & \multicolumn{3}{c}{Deletion given MSS correct} & \multicolumn{2}{c}{Addition} & Invalid \\
\cmidrule(lr){4-6}\cmidrule(lr){7-8}
 & invariant & & To $\bot$ & Original & Other & Robust & Drift & \\
\midrule
Claude Fable 5.1 (0-shot) & 47.9 & 26 & 15.4 & 57.7 & 26.9 & 41.7 & 12.5 & 0.0 \\
Claude Fable 5.1 (1-shot) & 27.1 & 23 & 4.3 & 60.9 & 34.8 & 37.5 & 10.4 & 0.0 \\
Claude Fable 5.1 (2-shot) & 22.9 & 27 & 22.2 & 48.1 & 29.6 & 37.5 & 18.8 & 0.0 \\
Claude Fable 5.1 (3-shot) & 25.0 & 23 & 17.4 & 56.5 & 26.1 & 39.6 & 8.3 & 0.0 \\
GPT-5.6 (0-shot) & 20.8 & 26 & 26.9 & 42.3 & 30.8 & 35.4 & 18.8 & 0.0 \\
GPT-5.6 (1-shot) & 18.8 & 28 & 28.6 & 35.7 & 35.7 & 41.7 & 16.7 & 0.0 \\
GPT-5.6 (2-shot) & 18.8 & 27 & 40.7 & 29.6 & 29.6 & 43.8 & 12.5 & 0.0 \\
GPT-5.6 (3-shot) & 12.5 & 30 & 33.3 & 23.3 & 43.3 & 47.9 & 14.6 & 0.0 \\
InternVL3-8B Base & 50.0 & 15 & 26.7 & 60.0 & 13.3 & 20.8 & 10.4 & 0.0 \\
InternVL3-8B LoRA & 68.8 & 11 & 36.4 & 36.4 & 27.3 & 8.3 & 14.6 & 0.0 \\
InternVL3.5-14B & 66.7 & 12 & 25.0 & 66.7 & 8.3 & 25.0 & 0.0 & 0.0 \\
Kimi-VL-A3B & 62.5 & 22 & 0.0 & 86.4 & 13.6 & 39.6 & 6.2 & 0.0 \\
LLaVA-OneVision-7B & 70.8 & 25 & 0.0 & 80.0 & 20.0 & 37.5 & 2.1 & 20.8 \\
MiniCPM-V 4.5 & 56.2 & 26 & 7.7 & 80.8 & 11.5 & 47.9 & 6.2 & 0.0 \\
Mistral Small 3.2 24B & 66.7 & 4 & 25.0 & 50.0 & 25.0 & 8.3 & 0.0 & 0.0 \\
Phi-4-Multimodal & 75.0 & 18 & 0.0 & 72.2 & 27.8 & 35.4 & 2.1 & 0.0 \\
Qwen2.5-VL-32B & 58.3 & 19 & 15.8 & 78.9 & 5.3 & 35.4 & 4.2 & 0.0 \\
Qwen2.5-VL-7B Base & 70.8 & 22 & 4.5 & 81.8 & 13.6 & 43.8 & 2.1 & 0.0 \\
Qwen2.5-VL-7B LoRA & 60.4 & 8 & 75.0 & 25.0 & 0.0 & 12.5 & 4.2 & 0.0 \\
Qwen3-VL-8B & 72.9 & 21 & 33.3 & 52.4 & 14.3 & 39.6 & 4.2 & 0.0 \\
\bottomrule
\end{tabular*}
\par\smallskip\parbox{\textwidth}{\footnotesize $m$ counts correct MSS answers out of 48 and is the deletion denominator. Other percentages use 48 groups. Original/Other denote spatial answers after deletion.}
\endgroup

\end{table*}

\subsection{Common initially correct groups}
\label{sec:shared-mss}
To separate response changes from changes in conditional-set membership,
we partition the 48 FQ2 groups by the primary Base and LoRA checkpoints'
MSS correctness. The both-correct, Base-only, LoRA-only and neither-correct
counts are 4/18/4/22 for Qwen and 9/6/2/31 for InternVL.
Table~\ref{tab:shared-mss} compares intervention answers on the common
groups and reconstructs each full joint difference. On those groups,
deletion has four gains and no losses for Qwen, versus one gain and no
losses for InternVL. Addition loses one Qwen and three InternVL successes,
with no gains.

\begin{table*}[!t]
  \centering
  \setlength{\abovecaptionskip}{0pt}
  \setlength{\belowcaptionskip}{5pt}
  \caption{Initial Base--LoRA comparisons on common MSS-correct groups.
  The full joint-success change equals the common-group change plus
  LoRA-only successes minus Base-only successes. All entries are counts.}
  \label{tab:shared-mss}

  \begingroup\footnotesize
  \setlength{\tabcolsep}{4pt}
  \begin{tabular*}{\textwidth}{@{\extracolsep{\fill}}llcrrrr@{}}
    \toprule
    Family & Metric & \shortstack{Common groups\\Base $\to$ LoRA}
      & \shortstack{Common\\change}
      & \shortstack{LoRA-only\\successes}
      & \shortstack{Base-only\\successes}
      & \shortstack{Full change\\out of 48} \\
    \midrule
    Qwen2.5-VL-7B & NTS & $0/4\to4/4$ & $+4$ & 2 & 1 & $+5$ \\
                  & DIS & $4/4\to3/4$ & $-1$ & 3 & 17 & $-15$ \\
    \midrule
    InternVL3-8B  & NTS & $2/9\to3/9$ & $+1$ & 1 & 2 & $0$ \\
                  & DIS & $7/9\to4/9$ & $-3$ & 0 & 3 & $-6$ \\
    \bottomrule
  \end{tabular*}
  \endgroup
  
\end{table*}

The remaining contribution comes from intervention successes in groups
whose MSS only one checkpoint answers correctly. Groups that neither
answers correctly contribute zero to both joint scores, whatever their
intervention answers. This exactly decomposes the original 48-group
numerators. The common sets are small and selected by observed outcomes,
so the paired changes describe these subsets rather than causal adaptation effects.
Reproduction materials retain group assignments and paired outcomes.

\subsection{Fixed-marginal joint-score diagnostic}
\label{sec:joint-marginals}
Low conjunction rates can follow from low component accuracy. Independent
uniform guesses give expected rates of $1/64$ for FQ1-GEM and $1/16$ for
NTS and DIS, but ignore the observed variant accuracies. We therefore add
a descriptive reference that keeps each variant's correct count fixed.
Let $C_{gj}$ indicate correctness for variant $j$ in group $g$, with
$c_j=\sum_g C_{gj}$ and $n=48$. The observed joint count is
$J=\sum_g\prod_{j=1}^{k}C_{gj}$. Fixing one correctness column and uniformly,
independently rearranging the others across groups gives the exact reference
rate $R=\prod_{j=1}^{k}(c_j/n)$. No sampled permutations are needed.
The possible joint count lies between
$\max(0,\sum_j c_j-(k-1)n)$ and $\min_j c_j$.

\begin{table*}[t]
  \centering
  \setlength{\abovecaptionskip}{0pt}
  \setlength{\belowcaptionskip}{5pt}
  \caption{Observed (O) and fixed-marginal rearrangement-reference (R)
  joint-success percentages for all primary configurations. Each criterion
  uses 48 groups. R preserves variant accuracy while independently
  rearranging answers across groups.}
  \label{tab:joint-marginals-primary}

  \begingroup\small
  \setlength{\tabcolsep}{6pt}
\begin{tabular*}{\textwidth}{@{\extracolsep{\fill}}lrrrrrr@{}}
\toprule
Configuration & \multicolumn{2}{c}{FQ1-GEM} & \multicolumn{2}{c}{NTS} & \multicolumn{2}{c}{DIS} \\
 & O & R & O & R & O & R \\
\midrule
Qwen2.5-VL-7B (Base) & 0.00 & 1.32 & 2.08 & 2.86 & 43.75 & 20.05 \\
InternVL3-8B (Base) & 2.08 & 1.88 & 8.33 & 10.42 & 20.83 & 8.46 \\
Qwen3-VL-8B (Base) & 0.00 & 0.80 & 14.58 & 20.96 & 39.58 & 17.32 \\
LLaVA-OneVision-7B (Base) & 0.00 & 1.11 & 0.00 & 1.09 & 37.50 & 19.53 \\
MiniCPM-V 4.5 (Base) & 0.00 & 0.98 & 4.17 & 6.77 & 47.92 & 28.21 \\
Mistral Small 3.2 24B (Base) & 0.00 & 1.56 & 2.08 & 4.17 & 8.33 & 1.74 \\
Qwen2.5-VL-32B (Base) & 0.00 & 1.14 & 6.25 & 14.84 & 35.42 & 17.32 \\
InternVL3.5-14B (Base) & 0.00 & 1.51 & 6.25 & 8.33 & 25.00 & 7.81 \\
Phi-4-Multimodal (Base) & 0.00 & 1.37 & 0.00 & 1.56 & 35.42 & 14.84 \\
Kimi-VL-A3B (Base) & 0.00 & 1.91 & 0.00 & 2.86 & 39.58 & 19.10 \\
Qwen2.5-VL-7B (LoRA) & 0.00 & 2.43 & 12.50 & 13.19 & 12.50 & 3.47 \\
InternVL3-8B (LoRA) & 0.00 & 2.79 & 8.33 & 14.32 & 8.33 & 3.82 \\
GPT-5.6 (0-shot) & 4.17 & 4.12 & 14.58 & 16.93 & 35.42 & 27.08 \\
GPT-5.6 (1-shot) & 6.25 & 5.53 & 16.67 & 17.01 & 41.67 & 35.24 \\
GPT-5.6 (2-shot) & 10.42 & 6.68 & 22.92 & 22.27 & 43.75 & 38.67 \\
GPT-5.6 (3-shot) & 6.25 & 8.23 & 20.83 & 23.44 & 47.92 & 40.36 \\
Claude Fable 5.1 (0-shot) & 0.00 & 2.40 & 8.33 & 11.28 & 41.67 & 31.60 \\
Claude Fable 5.1 (1-shot) & 6.25 & 6.79 & 2.08 & 4.99 & 37.50 & 23.96 \\
Claude Fable 5.1 (2-shot) & 6.25 & 3.26 & 12.50 & 17.58 & 37.50 & 28.12 \\
Claude Fable 5.1 (3-shot) & 2.08 & 4.48 & 8.33 & 14.97 & 39.58 & 24.96 \\
\bottomrule
\end{tabular*}

  \endgroup
  
\par\medskip
  \setlength{\abovecaptionskip}{0pt}
  \setlength{\belowcaptionskip}{5pt}
  \caption{The same joint-score diagnostic for the separate ordinary
  multi-seed cohort. O denotes observed rates and R the fixed-marginal
  reference. Each row retains all 48 groups per criterion.}
  \label{tab:joint-marginals-ordinary}

  \begingroup\small
  \setlength{\tabcolsep}{6pt}
\begin{tabular*}{\textwidth}{@{\extracolsep{\fill}}lrrrrrr@{}}
\toprule
Configuration & \multicolumn{2}{c}{FQ1-GEM} & \multicolumn{2}{c}{NTS} & \multicolumn{2}{c}{DIS} \\
 & O & R & O & R & O & R \\
\midrule
InternVL3-8B (matched control) & 2.08 & 2.28 & 8.33 & 11.11 & 20.83 & 8.33 \\
InternVL3-8B (LoRA 20260722) & 0.00 & 2.63 & 10.42 & 15.10 & 14.58 & 5.73 \\
InternVL3-8B (LoRA 20260723) & 0.00 & 2.45 & 8.33 & 13.37 & 10.42 & 3.82 \\
InternVL3-8B (LoRA 20260724) & 2.08 & 3.01 & 10.42 & 13.89 & 12.50 & 3.91 \\
Qwen2.5-VL-7B (matched control) & 0.00 & 1.19 & 2.08 & 2.86 & 45.83 & 21.01 \\
Qwen2.5-VL-7B (LoRA 20260722) & 0.00 & 2.63 & 8.33 & 11.55 & 12.50 & 3.04 \\
Qwen2.5-VL-7B (LoRA 20260723) & 0.00 & 2.43 & 8.33 & 11.55 & 12.50 & 3.04 \\
Qwen2.5-VL-7B (LoRA 20260724) & 0.00 & 3.42 & 14.58 & 15.62 & 12.50 & 4.77 \\
\bottomrule
\end{tabular*}

  \endgroup
  
\end{table*}

\begin{table*}[!t]
  \centering
  \setlength{\abovecaptionskip}{0pt}
  \setlength{\belowcaptionskip}{5pt}
  \caption{The fixed-marginal diagnostic applied to all three retained
  metadata controls. O is observed joint success and R the rearrangement
  reference. Rules are fitted on training groups. The constant-letter
  control produces opposite NTS and DIS gaps using answer preferences alone.}
  \label{tab:joint-control-reference}
\begingroup\small
  \setlength{\tabcolsep}{4pt}
  \begin{tabular*}{\textwidth}{@{\extracolsep{\fill}}lrrrrrrr@{}}
    \toprule
    Input features & \multicolumn{3}{c}{Correct counts / 48} &
      \multicolumn{2}{c}{NTS (\%)} & \multicolumn{2}{c}{DIS (\%)} \\
    \cmidrule(lr){2-4}\cmidrule(lr){5-6}\cmidrule(l){7-8}
      & MSS & Del. & Add. & O & R & O & R \\
    \midrule
    Majority letter & 11 & 12 & 11 & 0.00 & 5.73 & 22.92 & 5.25 \\
    Image count & 9 & 9 & 13 & 0.00 & 3.52 & 4.17 & 5.08 \\
    Count + insufficiency letter & 5 & 42 & 7 & 6.25 & 9.11 & 2.08 & 1.52 \\
    \bottomrule
  \end{tabular*}\endgroup
  
\end{table*}

Tables~\ref{tab:joint-marginals-primary} and
\ref{tab:joint-marginals-ordinary} retain all 20 primary and eight ordinary
configurations separately. Primary NTS is below the reference in 19 rows,
with gaps of $-8.59$ to $+0.65$ percentage points. DIS exceeds it in every
row, with gaps of $+4.51$ to $+23.70$ points. FQ1-GEM has five positive and
15 negative gaps, so it does not support a uniform direction. All eight
ordinary configurations have negative FQ1-GEM and NTS gaps and positive DIS
gaps. These configurations share test groups and are not independent
replications of a population-level effect.

The train-fitted controls in Table~\ref{tab:joint-control-reference}
make this limitation concrete. Always predicting the training
majority letter B gives NTS of 0.00\% against a 5.73\% reference and DIS of
22.92\% against a 5.25\% reference. Thus the same negative-NTS/positive-DIS
pattern can arise without images. This follows from the shared spatial answer
in the addition pair and its change to insufficiency after deletion.
Each control retains its train-fitted rule.

The reference breaks the observed pairing while preserving variant accuracy.
It therefore distinguishes how successes align from how often they occur.
However, heterogeneous groups are not exchangeable by design. Shared
difficulty and repeated answers can both affect the observed alignment.
We use the gaps and feasible bounds as descriptive measures of this alignment.
All groups and invalid outputs remain in the calculations. The companion
supplies component \mbox{counts and feasible bounds}.

\subsection{Uncertainty}
The paired contrasts in Table~\ref{tab:frameevidence-paired-contrasts}
resample exact-image connected components as
blocks while retaining paired outcomes within each contrast. It uses
10,000 bootstrap draws. These intervals condition on the held-out test and
exact-image grouping. They do not include provider nondeterminism,
demonstration-selection variance or all scene reuse. A zero-width interval at
zero indicates no observed group variation, not \mbox{zero population performance}.

\begin{table*}[!t]
  \centering
  \setlength{\abovecaptionskip}{0pt}
  \setlength{\belowcaptionskip}{5pt}
  \caption{Paired differences in percentage points, with 95\% cluster-block bootstrap intervals.}
\label{tab:frameevidence-paired-contrasts}

  \footnotesize
  \setlength{\tabcolsep}{1.6pt}
  \begin{tabular*}{\textwidth}{@{\extracolsep{\fill}}lrrrr@{}}
    \toprule
    Contrast & $\Delta$Acc. & $\Delta$FQ1-GEM & $\Delta$NTS & $\Delta$DIS \\
    \midrule
    InternVL3-8B : All--Text & -1.4 [-6.9,+4.2] & +0.0 [-5.9,+6.2] & +8.3 [+2.0,+17.0] & +12.5 [+0.0,+25.5] \\
    InternVL3-8B : All--One & +1.7 [-3.4,+6.7] & +2.1 [+0.0,+7.0] & +2.1 [-6.8,+11.5] & -8.3 [-20.0,+2.2] \\
    InternVL3-8B LoRA: All--Text & +2.4 [-3.5,+7.9] & +0.0 [+0.0,+0.0] & +8.3 [+2.0,+16.7] & +8.3 [+2.0,+16.7] \\
    InternVL3-8B LoRA: All--One & +4.9 [+1.3,+8.8] & +0.0 [+0.0,+0.0] & +4.2 [-3.7,+12.8] & -2.1 [-12.8,+8.3] \\
    InternVL3.5-14B : All--Text & -2.1 [-7.3,+3.1] & +0.0 [+0.0,+0.0] & +6.2 [+0.0,+13.3] & +10.4 [-2.1,+24.4] \\
    InternVL3.5-14B : All--One & +3.5 [-1.8,+8.7] & +0.0 [+0.0,+0.0] & -2.1 [-9.3,+4.5] & +16.7 [+4.3,+29.5] \\
    Kimi-VL-A3B : All--Text & +4.2 [-0.8,+9.3] & +0.0 [+0.0,+0.0] & +0.0 [+0.0,+0.0] & +4.2 [-7.0,+16.7] \\
    Kimi-VL-A3B : All--One & -1.0 [-6.6,+4.7] & -2.1 [-6.7,+0.0] & -6.2 [-13.0,+0.0] & +8.3 [-3.7,+22.4] \\
    LLaVA-OneVision-7B : All--Text & +0.0 [-5.3,+5.2] & +0.0 [+0.0,+0.0] & +0.0 [+0.0,+0.0] & -6.2 [-17.0,+4.1] \\
    LLaVA-OneVision-7B : All--One & -2.4 [-6.7,+1.5] & +0.0 [+0.0,+0.0] & +0.0 [+0.0,+0.0] & -12.5 [-23.8,-3.6] \\
    MiniCPM-V 4.5 : All--Text & +3.8 [-1.3,+9.7] & +0.0 [+0.0,+0.0] & +4.2 [+0.0,+10.2] & +14.6 [-2.0,+32.5] \\
    MiniCPM-V 4.5 : All--One & +1.0 [-3.8,+6.4] & +0.0 [+0.0,+0.0] & -4.2 [-12.5,+3.9] & +4.2 [-6.8,+16.3] \\
    Mistral Small 3.2 24B : All--Text & -3.8 [-7.1,-0.6] & +0.0 [+0.0,+0.0] & +2.1 [+0.0,+6.7] & +8.3 [+1.9,+17.1] \\
    Mistral Small 3.2 24B : All--One & -3.8 [-9.1,+2.0] & +0.0 [+0.0,+0.0] & -4.2 [-12.8,+4.1] & -22.9 [-35.6,-10.0] \\
    Phi-4-Multimodal : All--Text & -2.4 [-9.7,+4.9] & -2.1 [-6.7,+0.0] & +0.0 [+0.0,+0.0] & +16.7 [-4.2,+37.8] \\
    Phi-4-Multimodal : All--One & -1.7 [-6.7,+3.4] & +0.0 [+0.0,+0.0] & -4.2 [-10.0,+0.0] & -2.1 [-12.5,+8.7] \\
    Qwen2.5-VL-32B : All--Text & +2.4 [-4.7,+9.4] & +0.0 [+0.0,+0.0] & +6.2 [+0.0,+14.0] & +35.4 [+21.3,+51.0] \\
    Qwen2.5-VL-32B : All--One & +6.6 [+1.3,+11.9] & +0.0 [+0.0,+0.0] & -2.1 [-9.1,+4.8] & +18.8 [+3.9,+34.8] \\
    Qwen2.5-VL-7B : All--Text & -3.1 [-9.9,+3.9] & +0.0 [+0.0,+0.0] & +2.1 [+0.0,+6.8] & +43.8 [+29.2,+59.5] \\
    Qwen2.5-VL-7B : All--One & -0.3 [-3.3,+3.3] & +0.0 [+0.0,+0.0] & +0.0 [+0.0,+0.0] & +6.2 [-4.2,+17.8] \\
    Qwen2.5-VL-7B LoRA: All--Text & +3.1 [-2.0,+8.3] & -2.1 [-6.7,+0.0] & +12.5 [+4.2,+22.0] & +12.5 [+3.8,+23.4] \\
    Qwen2.5-VL-7B LoRA: All--One & +4.2 [-0.3,+8.9] & +0.0 [+0.0,+0.0] & +12.5 [+4.3,+22.2] & +12.5 [+3.8,+23.4] \\
    Qwen3-VL-8B : All--Text & +4.5 [-2.2,+11.4] & +0.0 [+0.0,+0.0] & +14.6 [+4.5,+26.2] & +33.3 [+18.2,+49.0] \\
    Qwen3-VL-8B : All--One & +5.2 [+0.0,+10.7] & +0.0 [+0.0,+0.0] & +6.2 [-4.8,+18.9] & +29.2 [+17.0,+42.0] \\
    Qwen2.5-VL-7B: LoRA--Base & +6.6 [-0.3,+12.8] & +0.0 [+0.0,+0.0] & +10.4 [+0.0,+20.9] & -31.2 [-51.2,-11.3] \\
    InternVL3-8B: LoRA--Base & +3.1 [-0.7,+7.2] & -2.1 [-7.0,+0.0] & +0.0 [-8.5,+8.2] & -12.5 [-22.9,-3.8] \\
    \bottomrule
  \end{tabular*}
  \par\smallskip
  \parbox{\textwidth}{\footnotesize Unmarked models are base checkpoints. Positive values favor the first condition. Paired outcome differences are retained while exact-image asset-component fallback clusters are resampled as blocks.}

\end{table*}

\subsection{Clustering sensitivity}
\label{sec:cluster-sensitivity}
\begin{table*}[!t]
\centering
  \setlength{\abovecaptionskip}{0pt}
  \setlength{\belowcaptionskip}{5pt}
  \caption{Post-hoc LoRA--Base sensitivity on 48 FQ2 groups. Changes and 95\% intervals are in percentage points. Gain/loss gives paired counts. Intervals use 20,000 block-bootstrap draws over 29 exact-image, 14 inferred-scene or 13 union components, with each draw normalized by its sampled group count.}
\label{tab:cluster-sensitivity}

\footnotesize
\setlength{\tabcolsep}{3pt}
\begin{tabular*}{\textwidth}{@{\extracolsep{\fill}}llrrrrr@{}}
\toprule
Family & Metric & $\Delta$ (pp) & Exact-image & Inferred-scene & Union & Gain/loss \\
\midrule
Qwen2.5-VL-7B & NTS & $+10.4$ & $[0.0, +20.9]$ & $[-2.6, +19.0]$ & $[-2.4, +19.2]$ & 6/1 \\
Qwen2.5-VL-7B & DIS & $-31.2$ & $[-51.1, -11.4]$ & $[-56.7, -8.7]$ & $[-57.1, -9.4]$ & 3/18 \\
\midrule
InternVL3-8B & NTS & $0.0$ & $[-8.5, +8.3]$ & $[-12.5, +8.2]$ & $[-12.5, +8.1]$ & 2/2 \\
InternVL3-8B & DIS & $-12.5$ & $[-22.8, -2.5]$ & $[-27.3, -2.5]$ & $[-27.3, -2.4]$ & 0/6 \\
\bottomrule
\end{tabular*}

\end{table*}

This post-hoc analysis compares four fixed LoRA--Base contrasts on 48 FQ2
groups. Table~\ref{tab:cluster-sensitivity} reports partitions into 29
exact-image components and 14 inferred core-MSS scene/template clusters.

For each partition, 20,000 bootstrap draws
(seed 20261009) sample its original number of clusters with replacement.
Each sampled block retains all its paired groups. A replicate divides its
signed success difference by its sampled group count, accounting for unequal
block sizes. The original point estimates always divide by 48.

These partitions are not nested. One exact-image component spanning groups
00360, 00361 and 00364 crosses two inferred scene clusters. We therefore also
join groups connected by either relation, including indirect paths. This
union yields 13 components, with at most 15 groups each. We apply the same
bootstrap scheme. Neither scene
labels nor union components certify independent physical scenes. Both NTS
intervals include zero, while both DIS intervals remain negative under all
three partitions.
InternVL's unchanged NTS also masks two newly successful and
\mbox{two newly failed groups}.

For a cluster with signed contribution $d_c$ and overall total $D$, we also
compute $100(D-d_c)/48$. This neutralizes its net contribution while holding
the full denominator fixed.
Across the inferred clusters, Qwen's NTS range remains positive and both DIS
ranges remain negative. The supplement includes per-group
outcomes, cluster contributions and a separate verification implementation.

\section{Deletion and Image-Reference Audit}
\label{sec:ordinal-audit}

Offline reconstruction reproduces all 864 question--mode inputs and 130
prepared images without position labels. The evaluator adds neither stable
original-image IDs nor missing-image placeholders. Both Circular input sets
preserve source questions and image order across 1,152 rotations each.
Reconstruction uses the saved implementation and prepared inputs rather than
archived wire requests.

\begin{table*}[!t]
\centering
  \setlength{\abovecaptionskip}{0pt}
  \setlength{\belowcaptionskip}{5pt}
  \caption{Deletion success across 48 groups stratified by the fate of the originally referenced image. $M$ counts correct MSS answers and $J$ jointly correct MSS/deletion answers. For a stratum of $n$ groups, NTS is $J/n$ and conditional deletion success is $J/M$ when $M>0$.}
\label{tab:ordinal-sensitivity}

\begingroup\footnotesize
\setlength{\tabcolsep}{4pt}
\begin{tabular*}{\textwidth}{@{\extracolsep{\fill}}llrrrrrrrr@{}}
\toprule
 & & \multicolumn{2}{c}{Removed ($n=36$)} & \multicolumn{2}{c}{Shifted ($n=6$)} & \multicolumn{2}{c}{Stable ($n=2$)} & \multicolumn{2}{c}{No ordinal ($n=4$)} \\
\cmidrule(lr){3-4}\cmidrule(lr){5-6}\cmidrule(lr){7-8}\cmidrule(l){9-10}
Model & Setting & $M$ & $J$ & $M$ & $J$ & $M$ & $J$ & $M$ & $J$ \\
\midrule
Claude Fable 5.1 & 0-shot & 22 & 4 & 2 & 0 & 2 & 0 & 0 & 0 \\
Claude Fable 5.1 & 1-shot & 19 & 0 & 1 & 1 & 2 & 0 & 1 & 0 \\
Claude Fable 5.1 & 2-shot & 21 & 4 & 2 & 1 & 1 & 0 & 3 & 1 \\
Claude Fable 5.1 & 3-shot & 19 & 3 & 2 & 1 & 1 & 0 & 1 & 0 \\
\midrule
GPT-5.6 & 0-shot & 20 & 5 & 3 & 1 & 2 & 1 & 1 & 0 \\
GPT-5.6 & 1-shot & 20 & 7 & 3 & 1 & 2 & 0 & 3 & 0 \\
GPT-5.6 & 2-shot & 20 & 8 & 5 & 3 & 0 & 0 & 2 & 0 \\
GPT-5.6 & 3-shot & 23 & 7 & 4 & 2 & 2 & 1 & 1 & 0 \\
\midrule
InternVL3-8B & Base & 14 & 4 & 1 & 0 & 0 & 0 & 0 & 0 \\
InternVL3-8B & LoRA & 10 & 4 & 1 & 0 & 0 & 0 & 0 & 0 \\
InternVL3.5-14B & Base & 12 & 3 & 0 & 0 & 0 & 0 & 0 & 0 \\
Kimi-VL-A3B & Base & 20 & 0 & 0 & 0 & 0 & 0 & 2 & 0 \\
LLaVA-OneVision-7B & Base & 19 & 0 & 2 & 0 & 1 & 0 & 3 & 0 \\
MiniCPM-V-4.5 & Base & 23 & 2 & 1 & 0 & 0 & 0 & 2 & 0 \\
Mistral-Small-3.2 & Base & 3 & 0 & 0 & 0 & 0 & 0 & 1 & 1 \\
Phi-4-Multimodal & Base & 16 & 0 & 2 & 0 & 0 & 0 & 0 & 0 \\
Qwen2.5-VL-32B & Base & 15 & 1 & 1 & 0 & 1 & 1 & 2 & 1 \\
Qwen2.5-VL-7B & Base & 20 & 1 & 0 & 0 & 1 & 0 & 1 & 0 \\
Qwen2.5-VL-7B & LoRA & 7 & 5 & 0 & 0 & 1 & 1 & 0 & 0 \\
Qwen3-VL-8B & Base & 19 & 7 & 1 & 0 & 0 & 0 & 1 & 0 \\
\bottomrule
\end{tabular*}
\endgroup

\end{table*}

\begin{table*}[!t]
  \centering
  \caption{GPT stable-ID prompting on the same twelve FQ2 triplets.
  Orig. and IDs count released-label matches. Changed counts answer
  differences between prompts in (a) and the left half of (b), and between
  blocks in the right half of (b).}
  \label{tab:stable-id-diagnostic}
  \footnotesize
  \setlength{\tabcolsep}{3pt}
  \begin{tabular*}{\textwidth}{@{\extracolsep{\fill}}lrrrr@{\hspace{20pt}}lrrrr@{}}
    \toprule
    \multicolumn{10}{@{}l}{\textit{(a) Counterbalanced prompt pairs}} \\
    Reference status & $n$ & Orig. & IDs & Changed & Input condition & $n$ & Orig. & IDs & Changed \\
    \midrule
    Retained, shifted & 18 & 12 & 8 & 8 & MSS & 12 & 8 & 6 & 6 \\
    Stable ordinal & 6 & 6 & 2 & 4 & Deletion & 12 & 7 & 4 & 6 \\
    No ordinal & 12 & 6 & 5 & 6 & Hard addition & 12 & 9 & 5 & 6 \\
    \midrule
    Overall & 36 & 24 & 15 & 18 & Overall & 36 & 24 & 15 & 18 \\
    \midrule
    \multicolumn{10}{@{}l}{\textit{(b) Two randomized execution blocks}} \\
    Block & $n$ & Orig. & IDs & Changed & Fixed prompt & $n$ & Block 1 & Block 2 & Changed \\
    \midrule
    1 & 36 & 17 & 17 & 12 & Original & 36 & 17 & 20 & 14 \\
    2 & 36 & 20 & 21 & 15 & Stable IDs & 36 & 17 & 21 & 15 \\
    \bottomrule
  \end{tabular*}
\end{table*}

The runner uses current image order. Comparing the first explicit Chinese
ordinal against the original and deleted lists identifies 36 removed
referents and six retained at changed positions. All 36 removed referents
are the original first image. Including groups 00368 and 00370 gives 38
ordinals pointing to another retained image. Groups 00360, 00361, 00364
and 00405 have out-of-range ordinals. Two retain the referent, while no
explicit ordinal is detected in four groups.

For example, group 00392 still asks for the first image's camera direction
after deleting that original image. Its request now begins with the original
second image. The question text is unchanged, but its ordinal now points to
a different image, coupling evidence removal with a reference change.

\paragraph{Sensitivity using the existing predictions.}
Table~\ref{tab:ordinal-sensitivity} links all 20 configurations to four
exhaustive strata without replacing the 48-group main denominator. Two
groups retain a stable explicit ordinal, and four have no detected ordinal.
For Qwen2.5-VL-7B, jointly correct deletion responses increase from one to
five among the 36 removed-anchor groups and from zero to one among the two
stable-ordinal groups. Both InternVL3-8B checkpoints obtain all four of their
joint successes in the removed-anchor stratum. All four checkpoints fail
jointly on the six retained-but-shifted groups.

Successes concentrate in removed-anchor groups rather than shifted-reference
cases. The six stable-ordinal or no-ordinal groups are too few to establish
the pattern independently.

\paragraph{Paired stable-ID prompting diagnostic.}
\label{sec:stable-id-diagnostic}
We select all twelve FQ2 groups outside the removed-anchor stratum,
covering six retained-but-shifted groups, two stable-ordinal groups and
four groups without an ordinal. Each group contributes MSS, deletion and
hard-addition inputs. The paired diagnostic obtains GPT responses
for both prompts, giving 36 items per arm and 72 valid responses. Images,
their order, options and released labels are held fixed. The rewrite
assigns stable original-image IDs, adds an explicit ID mapping and
instructions, and replaces ordinal references with those IDs. Original
wording is requested first on eighteen pairs and the rewrite first on
eighteen. Execution is interleaved rather than simultaneous.

Table~\ref{tab:stable-id-diagnostic}a reports released-label agreement and
prediction changes. Twelve pairs match the label in both arms, twelve
only under original wording, three only under the rewrite and nine in
neither. The overall agreement difference is $-25.00$ points. Paired
resampling of twelve groups gives a 95\% percentile interval of
[$-44.44$, $-5.56$] from 10,000 draws with seed 20261010. Using the six
image/scene union components gives [$-41.67$, $-3.03$] with seed 20261110.
The corresponding intervals for the 50.00\% prediction-change rate are
[30.56, 69.44]\% and [25.00, 70.83]\%.

\paragraph{Repeated-block control.}
The repeated-block control holds these 36 items and both prompt formulations
fixed across two blocks, each containing all 72 item--prompt combinations.
Within each block, requests are shuffled with seeds 20261012 and 20261013
and run with concurrency two. Block one finishes before block two starts.
This randomizes interleaving without enforcing which prompt is requested
first for each item. Images, their order, labels and decoding settings
remain unchanged. Only transport or HTTP failures are retried, at most
three attempts per request. All 144 responses are valid.

Across the blocks in Table~\ref{tab:stable-id-diagnostic}b, original wording
matches 37/72 labels and stable IDs match 38/72, a difference of 1.39 points.
A paired bootstrap with
20,000 draws over the twelve groups and seed 20261011 retains all three
conditions and both blocks within each group. Its 95\% interval is
[$-12.50$, $16.67$]. Resampling the six image/scene union components
instead gives [$-6.94$, $16.67$]. These intervals condition on the two
observed blocks. Across their 24 group--block observations,
NTS falls from 8 to 4 and DIS from 9 to 8, while three-answer joint
agreement remains 4. NTS has two paired gains and six losses despite
near-equal item agreement.

Original prompts change 14/36 answers between blocks, and stable-ID prompts
change 15/36. The initial 25-point decline therefore does not recur in the
repeated comparison. Response instability can coexist with nearly equal
aggregate agreement, so a single paired run does not isolate a wording
effect. The intervention combines IDs, mapping instructions and visible
numbering gaps rather than changing ID binding alone.

\paragraph{Targeted machine-assisted visual check.}
Given exact prompts and ordered images from six priority groups, two isolated
contexts of one assistant save answers before seeing gold labels. Both match
labels on the same 11 of 18 originals and abstain on seven owing to entity
or view ambiguity.
Audit abstention means inability to adjudicate, not an insufficient-information
answer. For nine additional single-view deletions, both select insufficiency
on six and abstain on three. Four original and five additional items have
out-of-range references that prevent judging necessity with unchanged
bindings. The saved responses and visual rationales localize entity and
reference ambiguity in this six-group model-assisted check.

\subsection{Original Inputs and Image-Order Traces}
\label{sec:input-traces}

\begin{figure*}[!t]
  \centering
  \includegraphics[width=\textwidth]{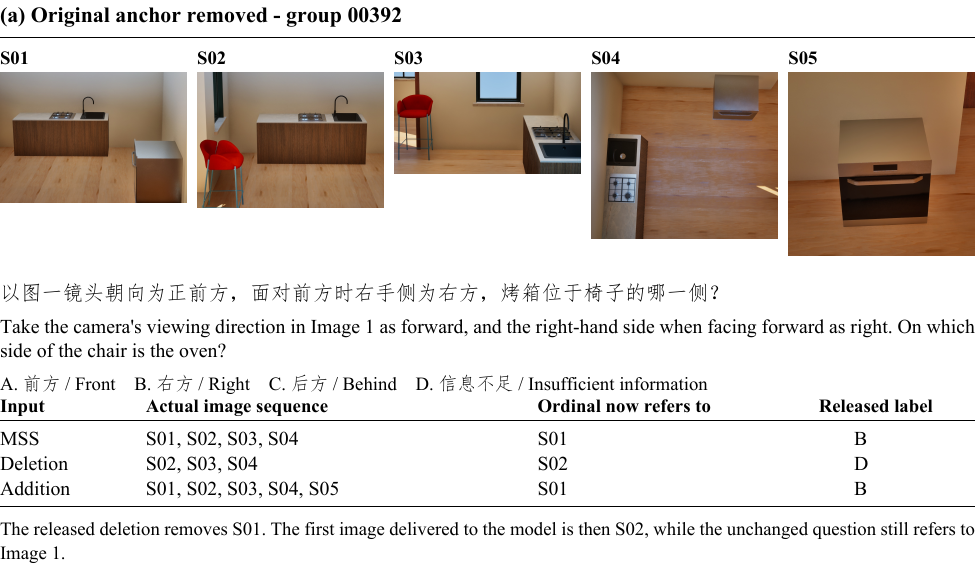}
  \par\vspace{9pt}
  \includegraphics[width=\textwidth]{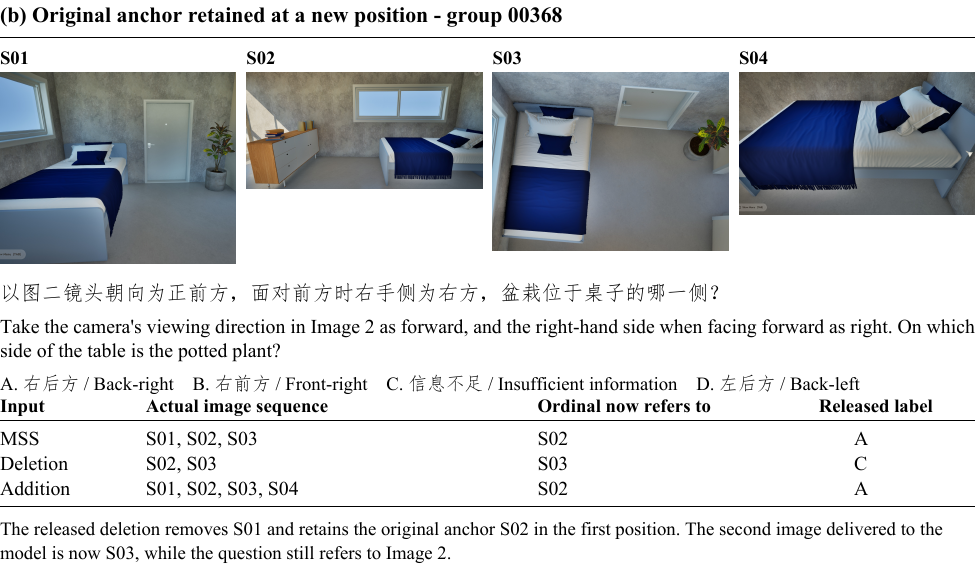}
  \caption{Original Chinese questions, options and image order in two FQ2
  triplets. English text translates the questions for reading. S identifiers
  track source images, while the original inputs use unlabeled current-order
  images. Labels are those of the released dataset.}
  \label{fig:original-input-traces}
\end{figure*}

Figure~\ref{fig:original-input-traces} reproduces two Chinese FQ2 triplets
with ordered images and English translations. Source IDs identify images
for comparison, while the evaluated requests use their current order.
The complete input audit covers all 48 test triplets and their 144 released
conditions. Removing one image preserves the order of the others but can
change which image an ordinal refers to. Group 00392 illustrates removal of
the original anchor, and group 00368 illustrates a retained anchor whose
position changes. Both retain the evaluated prompts and labels.

\section{Offline Diagnostic Supplements}
\ifdefined\FEinlineDiagnostic
\noindent\begin{minipage}{\textwidth}
\makeatletter\def\@captype{table}\makeatother
\else
\begin{table*}[!t]
\fi
  \centering
  \caption{Scene/template overlap sensitivity of the primary configurations.}
  \label{tab:scene-overlap-primary}
  \begingroup\small
  \setlength{\tabcolsep}{3pt}
  \renewcommand{\arraystretch}{1.14}
  \begin{tabular*}{\textwidth}{@{\extracolsep{\fill}}lrrrrrrrr@{}}
    \toprule
    & \multicolumn{4}{c}{Cluster shared with train} & \multicolumn{4}{c}{Cluster absent from train and dev} \\
    \cmidrule(lr){2-5}\cmidrule(l){6-9}
    Model / setting & Acc. & GEM & NTS & DIS & Acc. & GEM & NTS & DIS \\
    \midrule
    Qwen2.5-VL-7B & 28.40 & 0 & 1 & 17 & 31.11 & 0 & 0 & 4 \\
    InternVL3-8B & 27.57 & 1 & 3 & 6 & 40.00 & 0 & 1 & 4 \\
    Qwen3-VL-8B & 33.33 & 0 & 5 & 15 & 37.78 & 0 & 2 & 4 \\
    LLaVA-OneVision-7B & 29.22 & 0 & 0 & 16 & 28.89 & 0 & 0 & 2 \\
    MiniCPM-V 4.5 & 32.92 & 0 & 2 & 19 & 33.33 & 0 & 0 & 4 \\
    Mistral Small 3.2 24B & 25.10 & 0 & 1 & 4 & 28.89 & 0 & 0 & 0 \\
    Qwen2.5-VL-32B & 33.74 & 0 & 3 & 14 & 31.11 & 0 & 0 & 3 \\
    InternVL3.5-14B & 28.40 & 0 & 3 & 11 & 26.67 & 0 & 0 & 1 \\
    Phi-4-Multimodal & 26.34 & 0 & 0 & 14 & 26.67 & 0 & 0 & 3 \\
    Kimi-VL-A3B & 27.16 & 0 & 0 & 14 & 40.00 & 0 & 0 & 5 \\
    \midrule
    Qwen2.5-VL-7B / LoRA & 34.57 & 0 & 5 & 4 & 40.00 & 0 & 1 & 2 \\
    InternVL3-8B / LoRA & 30.45 & 0 & 2 & 3 & 44.44 & 0 & 2 & 1 \\
    \midrule
    GPT-5.6 / 0-shot & 41.56 & 2 & 6 & 15 & 35.56 & 0 & 1 & 2 \\
    \quad 1-shot & 42.39 & 3 & 6 & 16 & 53.33 & 0 & 2 & 4 \\
    \quad 2-shot & 46.91 & 5 & 6 & 15 & 55.56 & 0 & 5 & 6 \\
    \quad 3-shot & 51.03 & 3 & 8 & 19 & 44.44 & 0 & 2 & 4 \\
    \midrule
    Claude Fable 5.1 / 0-shot & 38.68 & 0 & 3 & 17 & 42.22 & 0 & 1 & 3 \\
    \quad 1-shot & 40.74 & 3 & 1 & 15 & 37.78 & 0 & 0 & 3 \\
    \quad 2-shot & 40.33 & 3 & 5 & 16 & 46.67 & 0 & 1 & 2 \\
    \quad 3-shot & 42.39 & 1 & 2 & 16 & 37.78 & 0 & 2 & 3 \\
    \bottomrule
  \end{tabular*}
  \par\smallskip
  \parbox{\textwidth}{\footnotesize\raggedright Acc. is a percentage over 243 or 45 questions. GEM is a count out of 41 or 7 FQ1 groups. NTS and DIS are counts out of 40 or 8 FQ2 groups. The partition uses an inferred scene/template ledger. No test cluster overlaps only dev. Invalid answers remain errors. Unmarked open models are bases.}
  \endgroup
\ifdefined\FEinlineDiagnostic
\end{minipage}
\else
\end{table*}
\fi

\subsection{Sensitivity to Scene/Template Overlap}
\label{sec:scene-overlap-supplement}

The inferred scene/template ledger and fixed split define
overlap with train, dev alone, or neither. These categories contain 81,
zero and 15 test groups, respectively. Rates are omitted for the empty
dev-only stratum.

The overlap stratum contains 41 FQ1 and 40 FQ2 groups across ten clusters.
The four non-overlapping clusters contain seven FQ1 and eight FQ2 groups.
Their 13 synthetic and two real groups contrast with 41 synthetic and
40 real overlap groups. These strata measure descriptive sensitivity
rather than a causal scene-exposure effect.

Table~\ref{tab:scene-overlap-primary} preserves all 20 primary configurations.
Every primary configuration has zero GEM on the seven non-overlapping FQ1 groups.
Qwen's primary LoRA reduces DIS in both strata, from 17/40 to 4/40 in
overlapping groups and from 4/8 to 2/8 elsewhere. InternVL's NTS changes
from 3/40 to 2/40 and from 1/8 to 2/8, respectively. The unchanged overall
NTS therefore combines opposing subgroup changes.
Table~\ref{tab:scene-overlap-ordinary} separately reports matched bases and
three training seeds.

Paired group outcomes further distinguish gains from losses.
Qwen's non-overlapping DIS decline comprises two gains and
four losses.
Across strata, numerators, denominators, gains and losses sum exactly to
their full-test counterparts. With eight FQ2 groups, each success changes
the non-overlapping rate by 12.5 points.

\ifdefined\FEinlineDiagnostic
\noindent\begin{minipage}{\textwidth}
\makeatletter\def\@captype{table}\makeatother
\else
\begin{table*}[!t]
\fi
  \centering
  \caption{Scene/template overlap sensitivity of the separate three-seed cohort.}
  \label{tab:scene-overlap-ordinary}
  \begingroup\footnotesize
  \setlength{\tabcolsep}{3pt}
  \renewcommand{\arraystretch}{1.02}
  \begin{tabular*}{\textwidth}{@{\extracolsep{\fill}}lrrrrrrrr@{}}
    \toprule
    & \multicolumn{4}{c}{Cluster shared with train} & \multicolumn{4}{c}{Cluster absent from train and dev} \\
    \cmidrule(lr){2-5}\cmidrule(l){6-9}
    Model / setting & Acc. & GEM & NTS & DIS & Acc. & GEM & NTS & DIS \\
    \midrule
    InternVL3-8B / Base control & 28.40 & 1 & 3 & 6 & 40.00 & 0 & 1 & 4 \\
    InternVL3-8B / LoRA 20260722 & 30.86 & 0 & 3 & 6 & 44.44 & 0 & 2 & 1 \\
    InternVL3-8B / LoRA 20260723 & 28.81 & 0 & 2 & 4 & 42.22 & 0 & 2 & 1 \\
    InternVL3-8B / LoRA 20260724 & 30.86 & 0 & 3 & 4 & 46.67 & 1 & 2 & 2 \\
    \midrule
    Qwen2.5-VL-7B / Base control & 28.81 & 0 & 1 & 18 & 31.11 & 0 & 0 & 4 \\
    Qwen2.5-VL-7B / LoRA 20260722 & 34.98 & 0 & 3 & 4 & 40.00 & 0 & 1 & 2 \\
    Qwen2.5-VL-7B / LoRA 20260723 & 33.74 & 0 & 3 & 4 & 40.00 & 0 & 1 & 2 \\
    Qwen2.5-VL-7B / LoRA 20260724 & 37.45 & 0 & 6 & 4 & 37.78 & 0 & 1 & 2 \\
    \bottomrule
  \end{tabular*}
  \par\smallskip
  \parbox{\textwidth}{\footnotesize\raggedright Acc. is a percentage over 243 or 45 questions. GEM is a count out of 41 or 7 FQ1 groups. NTS and DIS are counts out of 40 or 8 FQ2 groups. The partition uses an inferred scene/template ledger. No test cluster overlaps only dev. Seed numbers identify LoRA adapters, with a separate matched base for each family.}
  \endgroup
\ifdefined\FEinlineDiagnostic
\end{minipage}
\else
\end{table*}
\fi

\ifdefined\FEinlineDiagnostic
\par\noindent\begin{minipage}{\columnwidth}
\makeatletter\def\@captype{table}\makeatother
\else
\begin{table}[t]
\fi
  \centering
  \caption{FQ1 semantic changes across frame pairs.}
  \label{tab:fq1-pair-decomposition}
  \begingroup\footnotesize
  \setlength{\tabcolsep}{3pt}
  \renewcommand{\arraystretch}{1.02}
  \begin{tabular*}{\columnwidth}{@{\extracolsep{\fill}}lrrrr@{}}
    \toprule
    Model / setting & Same & Change & Both & GEM \\
    \midrule
    Qwen2.5-VL-7B & 114 & 30 & 4 & 0 \\
    InternVL3-8B & 94 & 50 & 4 & 1 \\
    Qwen3-VL-8B & 115 & 29 & 1 & 0 \\
    LLaVA-OneVision-7B & 116 & 28 & 4 & 0 \\
    MiniCPM-V 4.5 & 99 & 45 & 3 & 0 \\
    Mistral Small 3.2 24B & 112 & 32 & 2 & 0 \\
    Qwen2.5-VL-32B & 103 & 41 & 1 & 0 \\
    InternVL3.5-14B & 111 & 33 & 1 & 0 \\
    Phi-4-Multimodal & 120 & 24 & 1 & 0 \\
    Kimi-VL-A3B & 108 & 36 & 4 & 0 \\
    \midrule
    Qwen2.5-VL-7B / LoRA & 103 & 41 & 6 & 0 \\
    InternVL3-8B / LoRA & 113 & 31 & 6 & 0 \\
    \midrule
    GPT-5.6 / 0-shot & 54 & 90 & 18 & 2 \\
    \quad 1-shot & 54 & 90 & 21 & 3 \\
    \quad 2-shot & 50 & 94 & 26 & 5 \\
    \quad 3-shot & 50 & 94 & 27 & 3 \\
    \midrule
    Claude Fable 5.1 / 0-shot & 92 & 52 & 7 & 0 \\
    \quad 1-shot & 68 & 76 & 23 & 3 \\
    \quad 2-shot & 59 & 85 & 17 & 3 \\
    \quad 3-shot & 65 & 79 & 15 & 1 \\
    \bottomrule
  \end{tabular*}
  \par\smallskip
  \parbox{\columnwidth}{\footnotesize\raggedright Same and Change count predicted direction relations out of 144 frame pairs. Both counts pairs with two correct answers, a subset of Change. GEM counts fully correct triplets out of 48. All evaluated test gold pairs require a direction change and all responses are valid. Unmarked open models are bases.}
  \endgroup
\ifdefined\FEinlineDiagnostic
\end{minipage}\par\medskip
\else
\end{table}
\fi

\ifdefined\FEinlineDiagnostic
\par\noindent\begin{minipage}{\columnwidth}
\makeatletter\def\@captype{table}\makeatother
\else
\begin{table}[t]
\fi
  \centering
  \caption{Train-fitted question-and-option text baseline.}
  \label{tab:text-shortcut}
  \begingroup\footnotesize
  \setlength{\tabcolsep}{3pt}
  \renewcommand{\arraystretch}{1.02}
  \begin{tabular*}{\columnwidth}{@{\extracolsep{\fill}}lrr@{}}
    \toprule
    Metric & Dev (96 questions) & Test (288 questions) \\
    \midrule
    P & 25.29 & 23.99 \\
    R & 24.76 & 23.97 \\
    F1 & 24.63 & 23.97 \\
    Acc. & 25.00 & 23.96 \\
    \midrule
    GEM & 0/16 & 0/48 \\
    NTS & 0/16 & 0/48 \\
    DIS & 4/16 & 5/48 \\
    \bottomrule
  \end{tabular*}
  \par\smallskip
  \parbox{\columnwidth}{\footnotesize\raggedright P, R and F1 are macro-averaged percentages over A--D. GEM, NTS and DIS are joint-success counts. Character TF-IDF and logistic regression are fitted on 576 training questions with fixed settings. Dev does not guide model selection.}
  \endgroup
\ifdefined\FEinlineDiagnostic
\end{minipage}\par\medskip
\else
\end{table}
\fi

\subsection{Semantic Changes Across FQ1 Pairs}
\label{sec:fq1-pair-supplement}

Each of the 48 FQ1 triplets contributes camera--object, camera--world and
object--world pairs. Predicted letters are mapped to direction meanings.
The 144 pairs per configuration are partitioned by gold-meaning changes
and predicted changes, unchanged meanings or invalid responses.
Joint correctness is counted within each cell.

All FQ1 test gold pairs require a semantic change, and all predictions are
valid. False-change rates on unchanged gold therefore cannot be estimated.
The 2,880 configuration-pair records reuse the same 48 groups.

Table~\ref{tab:fq1-pair-decomposition} separates answer changes from successful
switching. GPT two- and three-shot each change 94 pairs, with 26 and 27 jointly
correct, yet fully correct triplets fall from five to three. Improved pairwise
correctness need not extend to all three frames. Qwen3 changes only 29 pairs,
with one jointly correct.

\subsection{Train-Only Text Shortcut Baseline}
\label{sec:offline-text}

We fit a single shallow classifier with fixed hyperparameters to the visible
question followed by its A--D option texts. Character TF-IDF uses 2--5-grams, minimum document frequency
two, a 50,000-feature cap, sublinear term frequency and L2 normalization,
without lowercasing.

Logistic regression uses $C=1$, the limited-memory Broyden--Fletcher--Goldfarb--Shanno
(L-BFGS) optimizer, no class
weighting, at most 5,000 iterations and seed 20261011.

Vocabulary, inverse document frequencies and classifier weights are learned
from the 576 training items. Identifiers, condition tags and image metadata
are excluded.

The fitted vocabulary contains 3,623 features, and optimization converges in
31 iterations without warnings. The 96 dev items provide a descriptive
check without tuning. One evaluation on the 288 test items gives 69 correct
answers. Table~\ref{tab:text-shortcut} reports macro classification metrics
and joint counts from the same saved predictions. Implementation uses
scikit-learn 1.7.2 on CPU.

The fixed text classifier performs near the four-choice accuracy reference
and obtains no fully correct FQ1 groups or MSS--deletion pairs. It nevertheless
solves five MSS--addition pairs, including all five groups on which its MSS
answer is correct. Addition agreement can thus persist even for a weak
text-only classifier.

\section{Additional Paired and Stratified Analyses}
\label{sec:source-crossrun}
\begin{table*}[!t]
  \begin{minipage}{\textwidth}
\makeatletter\def\@captype{table}\makeatother
\centering
\setlength{\abovecaptionskip}{0pt}
\setlength{\belowcaptionskip}{4pt}
\caption{All-images performance by image source (\%). Each source contains equal numbers of FQ1 and FQ2 groups.}
\label{tab:source-strata}
\begingroup\footnotesize
\setlength{\tabcolsep}{3pt}
\renewcommand{\arraystretch}{1.02}
\begin{tabular*}{\textwidth}{@{\extracolsep{\fill}}llrrrrrrrr@{}}
\toprule
 & & \multicolumn{4}{c}{Captured (126 questions)} & \multicolumn{4}{c}{Rendered (162 questions)} \\
\cmidrule(lr){3-6}\cmidrule(lr){7-10}
Model & Setting & Acc. & GEM & NTS & DIS & Acc. & GEM & NTS & DIS \\
\midrule
Qwen2.5-VL-7B & Base & 25.40 & 0.00 & 0.00 & 33.33 & 31.48 & 0.00 & 3.70 & 51.85 \\
InternVL3-8B & Base & 23.81 & 4.76 & 4.76 & 9.52 & 33.95 & 0.00 & 11.11 & 29.63 \\
Qwen3-VL-8B & Base & 32.54 & 0.00 & 9.52 & 33.33 & 35.19 & 0.00 & 18.52 & 44.44 \\
LLaVA-OneVision-7B & Base & 23.02 & 0.00 & 0.00 & 23.81 & 33.95 & 0.00 & 0.00 & 48.15 \\
MiniCPM-V 4.5 & Base & 30.16 & 0.00 & 4.76 & 33.33 & 35.19 & 0.00 & 3.70 & 59.26 \\
Mistral Small 3.2 24B & Base & 25.40 & 0.00 & 0.00 & 14.29 & 25.93 & 0.00 & 3.70 & 3.70 \\
Qwen2.5-VL-32B & Base & 34.92 & 0.00 & 9.52 & 38.10 & 32.10 & 0.00 & 3.70 & 33.33 \\
InternVL3.5-14B & Base & 28.57 & 0.00 & 4.76 & 19.05 & 27.78 & 0.00 & 7.41 & 29.63 \\
Phi-4-Multimodal & Base & 22.22 & 0.00 & 0.00 & 23.81 & 29.63 & 0.00 & 0.00 & 44.44 \\
Kimi-VL-A3B & Base & 23.02 & 0.00 & 0.00 & 23.81 & 33.95 & 0.00 & 0.00 & 51.85 \\
\midrule[0.3pt]
Qwen2.5-VL-7B & LoRA & 35.71 & 0.00 & 9.52 & 4.76 & 35.19 & 0.00 & 14.81 & 18.52 \\
InternVL3-8B & LoRA & 29.37 & 0.00 & 9.52 & 9.52 & 35.19 & 0.00 & 7.41 & 7.41 \\
\midrule[0.3pt]
GPT-5.6 & 0-shot & 43.65 & 9.52 & 14.29 & 38.10 & 38.27 & 0.00 & 14.81 & 33.33 \\
GPT-5.6 & 1-shot & 35.71 & 4.76 & 0.00 & 28.57 & 50.62 & 7.41 & 29.63 & 51.85 \\
GPT-5.6 & 2-shot & 48.41 & 19.05 & 9.52 & 28.57 & 48.15 & 3.70 & 33.33 & 55.56 \\
GPT-5.6 & 3-shot & 51.59 & 4.76 & 19.05 & 42.86 & 48.77 & 7.41 & 22.22 & 51.85 \\
\midrule[0.3pt]
Claude Fable 5.1 & 0-shot & 34.92 & 0.00 & 4.76 & 19.05 & 42.59 & 0.00 & 11.11 & 59.26 \\
Claude Fable 5.1 & 1-shot & 37.30 & 4.76 & 0.00 & 33.33 & 42.59 & 7.41 & 3.70 & 40.74 \\
Claude Fable 5.1 & 2-shot & 33.33 & 4.76 & 9.52 & 28.57 & 47.53 & 7.41 & 14.81 & 44.44 \\
Claude Fable 5.1 & 3-shot & 41.27 & 4.76 & 4.76 & 28.57 & 41.98 & 0.00 & 11.11 & 48.15 \\
\bottomrule
\end{tabular*}
\par\vspace{2pt}
\parbox{\textwidth}{\footnotesize\raggedright Joint metrics divide by 21 captured or 27 rendered groups. LLaVA retains six captured and four rendered invalid answers as errors. All rows reuse the same source strata.}
\endgroup
\end{minipage}

  \par\medskip
  \begin{minipage}{\textwidth}
\makeatletter\def\@captype{table}\makeatother
\centering
\setlength{\abovecaptionskip}{0pt}
\setlength{\belowcaptionskip}{4pt}
\caption{Paired Base executions on the same 288 questions. Correctness changes are relative to the primary run.}
\label{tab:base-crossrun}
\begingroup\footnotesize
\setlength{\tabcolsep}{3pt}
\renewcommand{\arraystretch}{1.02}
\begin{tabular*}{\textwidth}{@{\extracolsep{\fill}}lrrrrrr@{}}
\toprule
Model & Same answer & Agreement (\%) & Both correct & Primary only & Later only & Neither correct \\
\midrule
Qwen2.5-VL-7B & 281 & 97.57 & 82 & 1 & 2 & 203 \\
InternVL3-8B & 277 & 96.18 & 83 & 2 & 4 & 199 \\
\bottomrule
\end{tabular*}
\endgroup
\end{minipage}

\end{table*}

\begin{table*}[!t]
  \begin{minipage}{\textwidth}
\makeatletter\def\@captype{table}\makeatother
\setlength{\abovecaptionskip}{0pt}
\setlength{\belowcaptionskip}{4pt}
  \centering
  \caption{Adjacent-shot differences with paired component-bootstrap intervals. Item gains do not consistently extend to joint success.}
  \label{tab:shot-paired-intervals}
  \begingroup\footnotesize
  \setlength{\tabcolsep}{3pt}
  \renewcommand{\arraystretch}{1.03}
  \begin{tabular*}{\textwidth}{@{\extracolsep{\fill}}llrrrr@{}}
    \toprule
    Model & Shots & Acc. & GEM & NTS & DIS \\
    \midrule
    GPT & $0\to1$ & +3.47 [-3.83, +14.55] & +2.08 [-3.64, +11.11] & +2.08 [-10.45, +21.05] & +6.25 [-8.47, +29.03] \\
     & $1\to2$ & +4.17 [-2.86, +9.49] & +4.17 [-6.25, +10.87] & +6.25 [-9.38, +23.91] & +2.08 [-10.23, +23.33] \\
     & $2\to3$ & +1.74 [-4.90, +5.35] & -4.17 [-12.50, +8.33] & -2.08 [-19.51, +7.25] & +4.17 [-18.42, +19.72] \\
    \midrule
    Claude & $0\to1$ & +1.04 [-3.00, +6.33] & +6.25 [+0.00, +14.29] & -6.25 [-12.77, +0.00] & -4.17 [-24.14, +11.32] \\
     & $1\to2$ & +1.04 [-3.89, +8.77] & +0.00 [-6.06, +6.45] & +10.42 [+3.03, +16.67] & +0.00 [-10.20, +11.76] \\
     & $2\to3$ & +0.35 [-10.95, +7.14] & -4.17 [-10.26, +0.00] & -4.17 [-10.53, +5.26] & +2.08 [-9.38, +10.00] \\
    \bottomrule
  \end{tabular*}
  \par\vspace{2pt}
  \parbox{\textwidth}{\footnotesize\raggedright
  Entries are later-minus-earlier differences in percentage points with pointwise 95\% percentile intervals from 20,000 paired resamples of 13 whole components. Acc. uses 288 questions. Each joint metric uses 48 groups. The fixed-prediction intervals reflect test-component composition, not run-to-run variability.}
  \endgroup
\end{minipage}

  \par\medskip
  \begin{minipage}{\textwidth}
\makeatletter\def\@captype{table}\makeatother
\centering
\setlength{\abovecaptionskip}{0pt}
\setlength{\belowcaptionskip}{4pt}
\caption{Semantic answer retention after addition and insufficiency detection across FQ2.}
\label{tab:fq2-intervention-retention}
\begingroup\footnotesize
\setlength{\tabcolsep}{3pt}
\renewcommand{\arraystretch}{1.02}
\begin{tabular*}{\textwidth}{@{\extracolsep{\fill}}llrrrrrr@{}}
\toprule
 & & \multicolumn{3}{c}{All 48 MSS--addition pairs} & \shortstack{MSS-correct\\retention} & \multicolumn{2}{c}{Insufficiency (\%)} \\
\cmidrule(lr){3-5}\cmidrule(lr){7-8}
Model & Setting & Retained & Flipped & Invalid & Correct / eligible & P & R \\
\midrule
Qwen2.5-VL-7B & Base & 44 & 4 & 0 & 21/22 & 42.86 & 6.25 \\
InternVL3-8B & Base & 36 & 12 & 0 & 10/15 & 39.02 & 33.33 \\
Qwen3-VL-8B & Base & 45 & 3 & 0 & 19/21 & 62.16 & 47.92 \\
LLaVA-OneVision-7B & Base & 37 & 1 & 10 & 18/25 & 100.00 & 2.08 \\
MiniCPM-V 4.5 & Base & 42 & 6 & 0 & 23/26 & 100.00 & 12.50 \\
Mistral Small 3.2 24B & Base & 37 & 11 & 0 & 4/4 & 32.43 & 50.00 \\
Qwen2.5-VL-32B & Base & 41 & 7 & 0 & 17/19 & 51.43 & 37.50 \\
InternVL3.5-14B & Base & 42 & 6 & 0 & 12/12 & 44.44 & 33.33 \\
Phi-4-Multimodal & Base & 43 & 5 & 0 & 17/18 & 33.33 & 4.17 \\
Kimi-VL-A3B & Base & 42 & 6 & 0 & 19/22 & 50.00 & 6.25 \\
\midrule[0.3pt]
Qwen2.5-VL-7B & LoRA & 38 & 10 & 0 & 6/8 & 41.30 & 79.17 \\
InternVL3-8B & LoRA & 30 & 18 & 0 & 4/11 & 41.10 & 62.50 \\
\midrule[0.3pt]
GPT-5.6 & 0-shot & 28 & 20 & 0 & 17/26 & 83.33 & 31.25 \\
GPT-5.6 & 1-shot & 27 & 21 & 0 & 20/28 & 82.35 & 29.17 \\
GPT-5.6 & 2-shot & 28 & 20 & 0 & 21/27 & 86.36 & 39.58 \\
GPT-5.6 & 3-shot & 29 & 19 & 0 & 23/30 & 85.71 & 37.50 \\
\midrule[0.3pt]
Claude Fable 5.1 & 0-shot & 28 & 20 & 0 & 20/26 & 90.91 & 20.83 \\
Claude Fable 5.1 & 1-shot & 34 & 14 & 0 & 18/23 & 55.56 & 10.42 \\
Claude Fable 5.1 & 2-shot & 30 & 18 & 0 & 18/27 & 71.43 & 31.25 \\
Claude Fable 5.1 & 3-shot & 35 & 13 & 0 & 19/23 & 75.00 & 31.25 \\
\bottomrule
\end{tabular*}
\par\vspace{2pt}
\parbox{\textwidth}{\footnotesize\raggedright Retention compares decoded option meanings. The first three counts partition 48 pairs. Conditional retention divides by MSS-correct groups. Insufficiency precision and recall use 48 positive and 96 negative labels across all 144 FQ2 questions.}
\endgroup
\end{minipage}

\end{table*}

\subsection{Captured and rendered inputs}

Captured and rendered strata contain 42 and 54 test groups, respectively,
with 21 and 27 in each question family. Table~\ref{tab:source-strata}
retains all primary predictions and counts invalid responses as errors.

Rendered inputs yield higher accuracy in 14/20 configurations and higher
DIS in 16. Qwen2.5-VL-32B instead favors captured inputs on both endpoints,
unlike its 7B Base. GPT's accuracy advantage also shifts from captured to
rendered with one shot. These descriptive contrasts combine image-source,
scene and question differences.

\subsection{Paired Base executions}

Table~\ref{tab:base-crossrun} pairs 288 valid answers per Base run under
identical item, label, group and option mappings. Agreement compares the
selected answer letters.

Qwen changes seven answers but gains one net correct item, from 83 to 84.
InternVL changes eleven but gains two, from 85 to 87. Small accuracy
differences thus conceal answer-level changes. Qwen's DIS rises from 21/48
to 22/48, with unchanged GEM and NTS. InternVL's \mbox{joint counts stay unchanged}.

\subsection{Adjacent-shot uncertainty}
\label{sec:shot-uncertainty}

Table~\ref{tab:shot-paired-intervals} compares each adjacent shot setting
within both gateway families. We pair predictions on all 96 test groups
and connect groups through the union of the exact-image and inferred
scene/template maps. The resulting 13 components contain both question
families. Each of 20,000 bootstrap replicates samples 13 whole components
with replacement and recomputes the paired difference using its actual
question or group denominator. Seed 20261013 fixes the shared component draws.

The intervals are exploratory, pointwise 95\% percentile intervals. They
measure test-composition sensitivity conditional on the stored predictions,
separately from execution variability. GPT's two-to-three-shot change
adds five correct items but loses two complete frame groups. Both accuracy
and GEM intervals span zero.

\subsection{Semantic retention and insufficiency detection}
\label{sec:semantic-retention}

Table~\ref{tab:fq2-intervention-retention} separates repeated meanings from
correct retention after addition. We decode letters to option text before
comparing answers. Retained and flipped pairs require two valid predictions,
while invalid pairs remain separate. Conditional retention divides the DIS
numerator by all MSS-correct groups, including those with an invalid addition
response in the denominator.

Across 144 FQ2 questions, insufficiency detection has 48 positive deletion
labels and 96 negative MSS/addition labels. Invalid positive-label responses
count as missed positives, while invalid negative-label responses remain
errors \mbox{rather than true negatives}.

Qwen Base retains 21/22 initially correct answers after addition, compared
with 6/8 for LoRA. InternVL changes from 10/15 to 4/11. Joint DIS therefore
depends on both initial coverage and subsequent retention. Qwen LoRA also
produces 38 true and 54 false insufficiency responses, compared with three
and four for Base. Higher deletion recall thus comes with more false
insufficiency responses on the two answerable conditions.

\end{document}